%% file: KAN_CL.tex
\documentclass{article}

\usepackage{PRIMEarxiv}

\usepackage[utf8]{inputenc} 
\usepackage[T1]{fontenc}    
\usepackage{hyperref}       
\usepackage{url}            
\usepackage{booktabs}       
\usepackage{amsfonts}       
\usepackage{nicefrac}       
\usepackage{microtype}      
\usepackage{lipsum}
\usepackage{fancyhdr}       
\usepackage{graphicx}       
\graphicspath{{media/}}     
\usepackage{amsmath}
\usepackage{amssymb}
\usepackage{natbib}
\usepackage{dirtytalk}
\usepackage{xcolor}
\usepackage[toc,page]{appendix}
\usepackage{subcaption}
\usepackage{float}
\usepackage{authblk}
\title{A preliminary study on continual learning in computer vision using Kolmogorov-Arnold Networks 
}

\author[1]{Alessandro Cacciatore\thanks{a.cacciatore1@unimc.it}}
\author[2]{Valerio Morelli}
\author[2]{Federica Paganica}
\author[3]{Emanuele Frontoni}
\author[2]{Lucia Migliorelli}
\author[2]{Daniele Berardini}
\affil[1]{Department of Humanities – Languages, Language Liaison, History, Arts, Philosophy, University of Macerata}
\affil[2]{Department of Information Engineering, Università Politecnica delle Marche}
\affil[3]{Department of Political Sciences, Communication and International Relations, University of Macerata}

\begin{document}
\maketitle

\begin{abstract}
Deep learning has long been dominated by multi-layer perceptrons (MLPs), which have demonstrated superiority over other optimizable models in various domains. Recently, a new alternative to MLPs has emerged —Kolmogorov-Arnold Networks (KAN)— which are based on a fundamentally different mathematical framework. According to their authors, KANs address several major issues in MLPs, such as catastrophic forgetting in continual learning scenarios. However, this claim has only been supported by results from a regression task on a toy 1D dataset. In this paper, we extend the investigation by evaluating the performance of KANs in continual learning tasks within computer vision, specifically using the MNIST datasets. To this end, we conduct a structured analysis of the behavior of MLPs and two KAN-based models in a class-incremental learning scenario, ensuring that the architectures involved have the same number of trainable parameters. Our results demonstrate that an efficient version of KAN outperforms both traditional MLPs and the original KAN implementation. We further analyze the influence of hyperparameters in MLPs and KANs, as well as the impact of certain trainable parameters in KANs, such as bias and scale weights. Additionally, we provide a preliminary investigation of recent KAN-based convolutional networks and compare their performance with that of traditional convolutional neural networks. Our codes can be found at \href{https://github.com/MrPio/KAN-Continual_Learning_tests}{https://github.com/MrPio/KAN-Continual\_Learning\_tests}.
\end{abstract}


\input{1intro}
\input{2sota}

\input{3KANarch}
\input{4results}

\input{5discussion}
\input{6conclusions}

\bibliographystyle{plainnat}
\bibliography{references}  

\newpage
\begin{appendices}
\input{9appendix}
\end{appendices}

\end{document}

%% file: 1intro.tex
\section{Introduction to Kolmogorov-Arnold Networks}
\label{sec:intro}

On 30 April 2024, \cite{liu2024kan} published their work on Kolmogorov-Arnold networks (KAN), a novel deep optimizable modelthat is fundamentally different from multi-layer perceptrons (MLP) in terms of architecture and mathematical foundations. In particular, MLPs are based on the universal approximation theorem (UAT), which states that an artificial neural network can approximate any function, provided it is sufficiently deep. In this sense, the theorem demonstrates the universality of neural architectures, i.e., their ability to model any function \citep{hornik1989multilayer}. MLPs consist of neurons — computational units that apply multiplication and addition operations to the input variables using trainable parameters (weights and biases). These neurons are organized into multiple layers, which process input data sequentially. MLPs form the foundation of deep learning models, and are widely used in various domains, including computer vision and natural language processing.

\subsection{The Kolmogorov-Arnold Theorem}
\label{subsec:KAT}
Unlike MLPs, KANs are based on the Kolmogorov-Arnold Theorem (KAT), which states that every multivariate continuous function $f:[0,1]^n\xrightarrow{} \mathbb{R}$, with $n\ge2$, can be expressed as a superposition of continuous functions of one variable, combined through two-argument addition:

\[f(\vec{x})=\sum_{q=0}^{2n}\chi_q\left(\sum_{p=1}^{n}\psi_{p,q}(x_p)\right)\]

where $\psi_{p,q}:[0,1]\xrightarrow{} \mathbb{R}$ are called \textit{inner functions} and $\chi_{q}:\mathbb{R}\xrightarrow{} \mathbb{R}$ are the \textit{outer functions}. Inner and outer functions are central to the optimization process in KANs. Unlike MLPs, which optimize individual weights, KANs optimize entire functions to model the relationships between inputs and outputs. More specifically, \textbf{KANs replace linear weights on edges with univariate functions}.

Over the years, several variants of the theorem have been proposed. In 1962, George Lorentz proved that the $2n+1$ outer functions $\chi_q$ can be replaced by a single function $\chi$ \citep{lorentz1962metric}. In 1965, David Sprecher showed that the $n(2n+1)$ inner functions $\psi_{p,q}$ can be replaced by a single function $\psi:[0,1]\xrightarrow{}[0,1]$ with an appropriate shift in its argument, given by one scalar $\eta$, and with a proper scaling factor given by $n$ real values $\lambda_1,...,\lambda_n$ \citep{sprecher1965structure}. 

In its basic form, the KAT shares similarities with the mathematical formulation of the UAT:

\[f(\vec{x})=\sum_{q=0}^{\infty}\omega_q\cdot\sigma\left(\sum_{p=1}^{n}\omega_{p,q}\cdot x_p\right)\]

where $\omega_q$ and $\omega_{p,q}$ are trainable parameters and $\sigma$ is a fixed activation function, necessary to introduce non-linearity in MLPs. There are two main differences between the two theorems:
\begin{itemize}
    \item While the KAT guarantees that $f(\vec{x})$ can be exactly represented using $n\times(2n+1)$ sums of univarate functions, the UAT provides only an approximation of $f$, as it requires an infinite number of outer functions;
    \item The UAT requires pre-defined activation functions ($\sigma$) to introduce non-linearity in the multivariate function, which could otherwise only model linear problems due to its reliance on linear combinations of inputs and weights. The KAT, in contrast, spreads the non-linearity across all inner and outer functions, which are not constrained a priori in their definition and can take any shape or degree.
\end{itemize}

This second point is the main reason why the KAT was never applied to machine learning before: inner and outer functions could assume any shape, possibly even non-smooth or fractal, which would make their optimization hardly possible \citep{liu2024kan}.


\subsection{Kolmogorov-Arnold Networks}
\label{subsec:KAN}
\begin{figure}[h]
	\centering
	\includegraphics [width=0.45\columnwidth, angle=0]{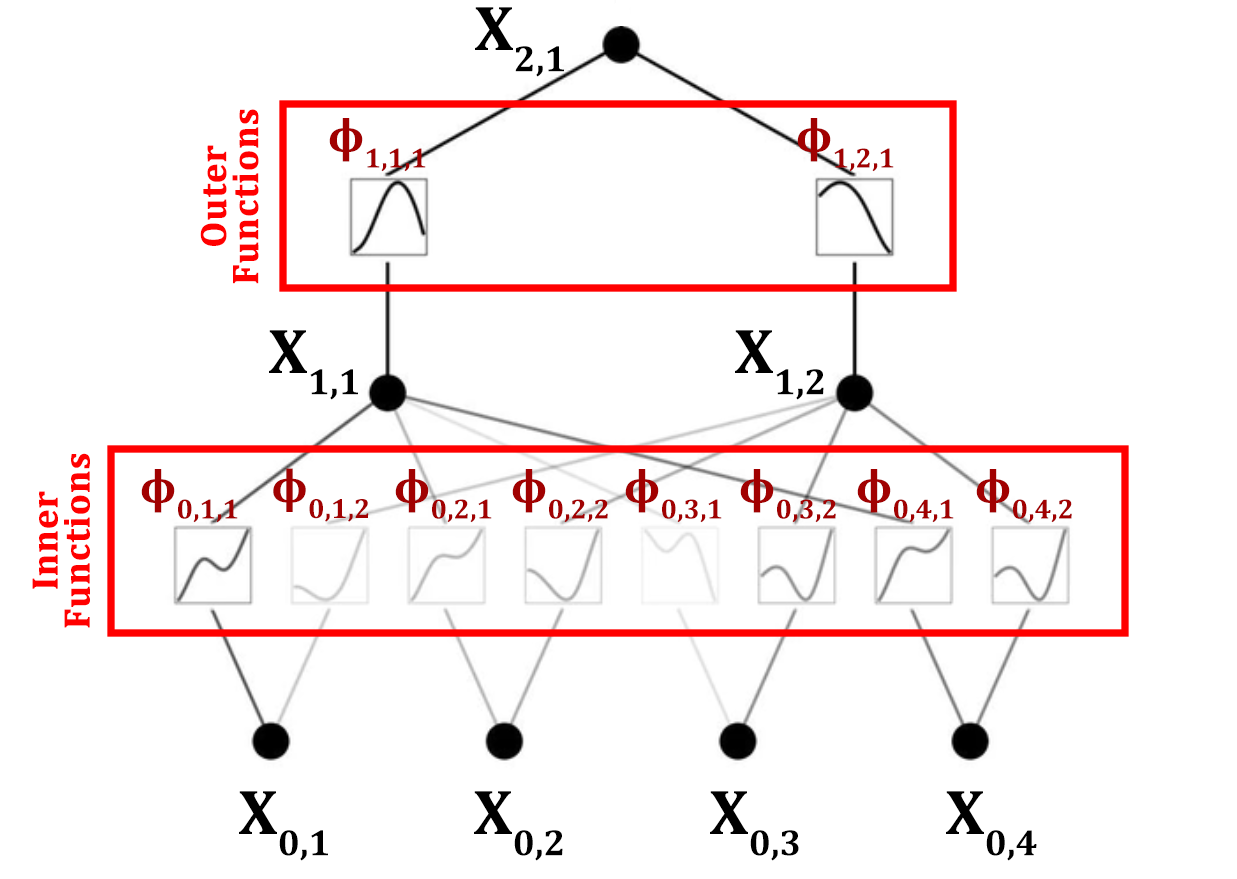}
	\caption{A simple two-layer KAN, where the activation functions are arranged on the edges and the nodes compute the sum. The number of outer functions does not comply with the KAT statement, since they are less than $2n+1=9$.}
	\label{fig:2_layer_kan}
\end{figure}

MLPs practically implement the UAT, and to train an MLP is to learn the optimal set of $\omega_q$ and $\omega_{p,q}$ for the problem at hand. KANs implement the KAT, and training a KAN requires optimizing both its outer ($\chi_q$) and inner ($\psi_{p,q}$) functions. As explained by the authors of the original paper, MLPs have fixed activation functions on nodes (“neurons”), whereas KANs feature learnable activation functions on edges (“weights”) \citep{liu2024kan}. It should be noted, however, that, contrary to what is stated in the paper, MLPs can also have learnable activation functions, such as \textit{parametric ReLU}, \textit{Swish}, and the more recent \textit{Maxout}. Furthermore, \cite{dhiman2024kan} points out that, if the activation function in an MLP is defined as $\phi(x)=\sigma(W x)$, then it can be formally considered a learnable activation function.

As can be seen in Fig. \ref{fig:2_layer_kan}, the architecture of a KAN resembles that of an MLP, with the key difference being that hidden and output nodes in a KAN ($X_{1,1}$, $X_{1,2}$, and $X_{2,1}$) are the summation of the results from the trainable functions located on the edges of the network. The inner and outer functions ($\chi_q$ and $\psi_{p,q}$) are not polynomial, as polynomials can exhibit a potentially \say{wild} and unpredictable behavior when optimized to fit a set of points. Specifically, as the number of points increases, polynomials tend to show abrupt changes in slope, which increases the risk of overfitting the training data. Moreover, polynomials offer low locality control since any part of the curve is highly dependent on all training samples. Instead, the authors \citep{liu2024kan} choose to parametrize inner and outer functions using \textit{B-splines}. This decision represents their key contribution, as KANs are not the first neural networks to be inspired by the KAT, and splines have been previously employed as learnable activation functions \citep{bohra2020learning}. A B-spline relies on non-learnable basis functions of fixed order ($k$) that are defined only over sub-intervals of the function domain, and the overall curve is obtained by joining these locally-defined functions in points called \say{knots}. The number of intervals in which the input domain is divided is referred to as the grid size ($G$). Together, $G$ and $k$ determine the number of basis functions used to map the input domain. Since the basis functions are defined a priori, the optimization process only involves the coefficients that scale each of these functions. Formally, this can be seen as:

\begin{equation}
  spline(x)=\sum_{i=1}^{G+k}c_iB_i(x),  
  \label{eq:spline_x}
\end{equation}

where $c_i$ are the coefficients that multiply the pre-defined basis functions ($B_i$). It is important to note that $spline(x)$ is not the final definition of the activation functions in KANs: further measures are taken by the authors in \cite{liu2024kan} in order to facilitate KANs optimization, as described in Sec. \ref{subsec:KANimplemCL}.

In principle, each $B_i(x)$ does not depend on the others, and can, therefore, more accurately fit the local training samples. B-splines are by definition $C^2$, meaning they are smooth enough to ensure the existence of a continuous second-order derivative. This property allows the network to be trained via backpropagation, while constraining the degrees of freedom of the local basis functions to adapt to local training samples. As demonstrated by \cite{de1978practical}, the KAT still holds when B-splines are used; however, in this case, the theorem guarantees an approximation of the original function rather than its exact representation. Consequently, some error is expected from this optimization process. Nevertheless \cite{liu2024kan} and \cite{de1978practical} show that approximation error has an upper bound with negligible dependence on input dimensions, suggesting that KANs are well-suited for modeling functions without suffering from the curse of dimensionality.\footnote{The original KAN paper \citep{liu2024kan} reports that test loss is inversely proportional to number of parameters N, roughly following the relationship $l \propto N^{-4}$. However, this relationship holds primarily for problems with a small number of input variables. Specifically, Fig. 3.1 indicates that test loss saturates with 100-dimensional input data.}. Furthermore, the authors overcome the limitations of the KAT --- which, in its basic form, corresponds to a two-layer network ---  by optimizing KANs with arbitrary number of layers and widths. 
In summary, the actual implementation of KANs relies on a relaxed version of the KAT, meaning they no longer exactly represent the multivariate function but, like MLPs, approximate the function underlying the data distribution.

\cite{liu2024kan} show that KANs possess desirable properties that make them a strong competitor to MLP. For instance, while a KAN is significantly more computationally intensive than an MLP with the same architecture (i.e., same depth and width), KANs typically require far fewer parameters (layers and edges) to achieve comparable performance. Additionally, KANs are naturally interpretable as they directly learn functions, and can be easily pruned by incorporating a sparsity penalty into the optimization process, allowing for the computation of the relevance of an edge (or activation function) to the processing carried out by the whole architecture. Another important feature of KANs highlighted by the authors is their remarkable behavior in continual learning (CL) scenarios, particularly regarding catastrophic forgetting, i.e., the tendency of artificial neural networks to forget previous knowledge when retrained on new tasks. Although the authors in \cite{liu2024kan} assert that KANs outperform MLPs in CL settings, they only back this claim with the results from a toy regression dataset, leaving further analysis for future work. Consequently, this paper presents a preliminary investigation into the CL ability of KANs (and KAN-based architectures) in more complicated tasks, namely multi-class image classification on the MNIST dataset \citep{lecun1998gradient}. To this aim, KANs are compared to MLPs in terms of their ability to store and learn new knowledge in a class-incremental learning task.

\subsection{Practical KAN implementation for continual learning}
\label{subsec:KANimplemCL}

In the original KAN framework (PyKAN \citep{pyKAN}), the authors employ several strategies to enhance the optimizability of KANs. Among these strategies, they include a bias and two scale factors for each activation function. The actual implementation of an activation function in a KAN layer is:

\begin{equation}
    \phi(x) = w_b b(x) + w_s spline(x) + \beta
    \label{eq:output_x}
\end{equation}

where:
\begin{itemize}
    \item $b(x)$ is a residual, non-optimizable SiLU activation function applied to the input variable;
    \item $w_b$ is a scale factor for the fixed activation $b(x)$;
    \item $spline(x)$ is defined in Eq. \ref{eq:spline_x};
    \item $w_s$ is a scale factor for the trainable activation $spline(x)$;
    \item $\beta$ is a bias factor that shifts the output.
\end{itemize}

Non-trainable $b(x)$ functions are not shown in Fig. \ref{fig:2_layer_kan}, and they should be considered as edges that run parallel to those connecting a node to its spline functions.

\begin{figure}
    \centering
    \includegraphics[width=0.9\linewidth]{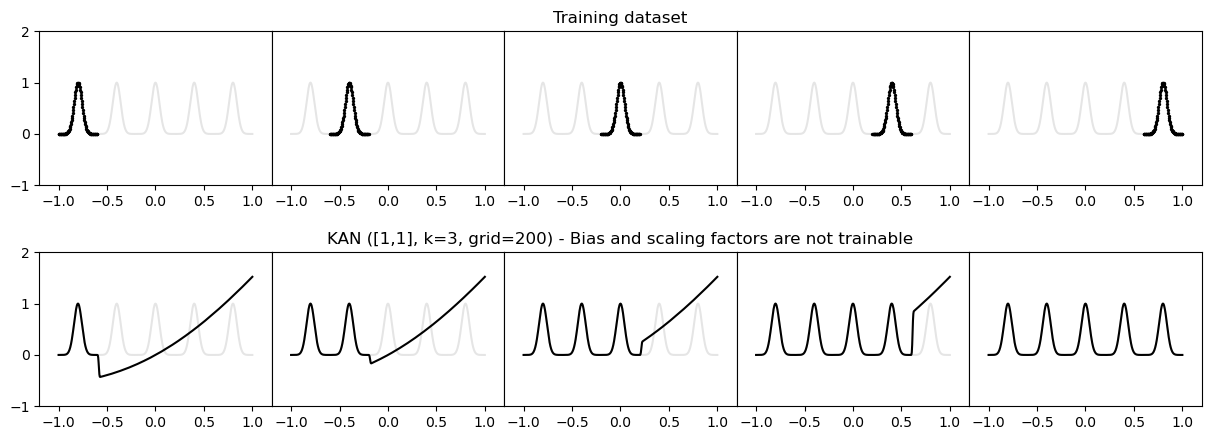}
    \caption{Toy example of KAN's ability in CL scenarios. The model is trained on a 1D regression dataset, by feeding points from each peaks sequentially. A [1,1] KAN with grid size set to 200 and spline order to 3 can perfectly fit data points, and new data points do not seem to have any influence on previously learnt knowledge.}
    \label{fig:CL_toy_dataset}
\end{figure}

As explained in the previous sections, B-splines rely on pre-defined basis functions, but the ability to scale or shift them improves their capability to fit a set of data points. For this reason, the bias $\beta$ and the two scale factors ($w_s$ and $w_b$) are optimized during training, along with the parameters that multiply each spline function, to allow the splines to adapt optimally to the training data. Unlike the trainable parameters of spline functions ($c_1,...,c_{G+K}$), the scale and bias parameters affect the overall spline-based activation rather than individual basis functions. As a result, bias and scale parameters do not contribute to—and may even hinder— the ability of splines to locally fit data points without being influenced by points outside their domain of interest. This issue is particularly problematic in CL scenarios, where the order in which training data is presented to the model is not random, and may depend on data labels (see Sec. \ref{subsec:CLscenarios}). In such contexts, data from a given task may exhibit different distributions or belong to different domains compared to data from previous tasks on which the model was trained.
To mitigate the impact of random or out-of-domain activation function, the authors find it beneficial to make these parameters non-trainable in the CL task they propose. To evaluate this, we replicated the experiment conducted by \cite{liu2024kan} in any possible configuration based on whether the bias and scaling factors were set as trainable parameters. In particular, the experiment involves a [1, 1] KAN (grid size set to 200 and spline order of 3) trained on a toy 1D regression dataset consisting of five consecutive Gaussian peaks. The five peaks are presented as five separate tasks, with data points from each peak sequentially fed to the KAN. As shown in Fig. \ref{fig:CL_toy_dataset}, the KAN model perfectly fits the training data without exhibiting any CL-related forgetting when new training points are introduced. Appendix \ref{app:PyvsEff} includes an in-depth study of KAN's ability to deal with CL scenarios depending on whether the bias and scaling factors are fixed or optimized during training. This study suggests that making these parameters non-optimizable is indeed the best possible option in 1D regression CL settings. Nevertheless, as explained in Sec. \ref{subsec:architectures}, this choice is not beneficial in multi-dimensional cases.


\subsection{KAN-based neural networks: EfficientKAN}
\label{subsec:EffKAN}

Since the seminal work on KANs \citep{liu2024kan} and the related code \citep{pyKAN} were released, numerous applications of KANs in neural networks have been developed and made publicly available, collected in a GitHub repository \citep{awesomeKAN}. In particular, we focus on \textit{EfficientKAN} (EffKAN) \citep{efficient-kan}, an optimized implementation of the original PyKAN framework \citep{pyKAN}. In Appendix \ref{app:ConvArchit} we additionally overview some of the proposed KAN-based convolutional neural networks.

The original PyKAN expands the input variables and stores all the intermediate variables involved in the \texttt{forward} pass. As illustrated in Fig. \ref{fig:2_layer_kan}, this involves expanding each input $x_{0,i}$ from a scalar variable to a (2, 1) tensor and storing it (\textit{pre-activations}), storing each $\Phi_{0,i,j}$ (\textit{post-spline values}), as well as storing the total activations (\textit{post-activations}, $\phi(x)$ in Eq. \ref{eq:output_x}). Expanding inputs to tensors and storing all pre- and post-activations requires significant memory, yet none of these operations is necessary for the \texttt{forward} pass. Pre-activations and post-activations are used to plot and prune a trained KAN, as well as to find the symbolic representation of the learned activation functions. 
EffKAN reformulates the computation: activation functions are applied directly as a scalar product of the B-splines $B_1(x),...,B_{G+k}(x)$ over the spline coefficients $c_1,...,c_{G+K}$ using the PyTorch \texttt{F.linear} layer. This significantly reduces memory costs and makes the computation a matrix multiplication operation, which is efficient for both forward and backward passes. 

However, the increased efficiency comes at a cost in terms of regularization. In fact, PyKAN employs L1 regularization on spline functions based on input samples to ensure sparisification during training. This operation is incompatible with the matrix-based reformulation of EffKAN, which instead uses an L1-regularization on the spline weights ($c_1,...,c_{G+K}$, i.e., the scalar coefficients that multiply each basis function). Moreover, EffKAN does not directly allow for pruning a trained KAN or regressing its symbolic representation, which is the mathematical expression described by each spline function. 

EffKAN implements $w_b$, $w_s$ (the scale factors in Eq. \ref{eq:output_x}) but not $\beta$. These parameters are further investigated in Appendix \ref{app:PyvsEff}, where we demonstrate that PyKAN and EffKAN are directly comparable in simple CL scenarios, such as the 1D Gaussian peaks regression.

%% file: 2sota.tex
\section{State of the art}
\label{sec:sota}

\subsection{State of the art on continual learning}
Despite the impressive results achieved by MLPs in across various tasks, one of their significant limitations is the  stark contrast to human intelligence in learning more tasks sequentially. CL is the setting in which models are updated incrementally as new data become available, rather than being trained on a fixed dataset. The ability to adapt to new data is essential for applications that require real-time adaptation to evolving information. However, this ability is compromised by a major drawback of current artificial neural networks: catastrophic forgetting. This phenomenon occurs when MLPs \say{forget} previously acquired knowledge when integrating new information. More specifically, MLP are subject to the stability-plasticity dilemma \citep{abraham2005memory}, because of which they can either display strong learning plasticity or strong memory stability. Plasticity allows them to adapt to new tasks, while stability enables them to retain knowledge from previous tasks. From an optimization point of view, plasticity requires weights adaptation via back-propagation, whereas stability is only possible insofar as the weight configuration of the network does not change. 

Liu and colleagues \citep{liu2024kan} argue that KANs are inherently suited for CL, showing catastrophic forgetting issues to a lesser extents. The authors attribute this favorable behavior to the property of \textit{local plasticity} that KANs inherit from splines. When fitting a dataset with a spline, single dataset points influence only the local spline coefficients (as discussed in Sec. \ref{subsec:KAN}), meaning that new data points will affect the local spline only if they belong to that specific domain. In CL settings, new data are often differ from the data encountered in previous tasks: consequently, they will be fitted by local functions that were likely not involved in earlier optimizations. Since KANs rely on splines, the whole architecture will benefit from the locality of the single activation functions that define it. In contrast, since MLPs use global activations, any local change may propagate uncontrollably across distant regions of the input domain, potentially disrupting the information stored in those intervals. In summary, spline optimization is confined to specific intervals of the curve domain, making it more precise than MLP optimization. As a result, when trained to fit simple datasets, KANs do not suffer from the stability-plasticity dilemma, because optimization is locally tailored to each data point, and information is safely stored by single splines. 

However, this might not be the case on more complex scenarios. The authors' evaluation \citep{liu2024kan} is limited to a simple 1D regression task, validated with a toy example involving sequential data around five Gaussian peaks. We intend to further enrich the investigation into KAN's ability to handle catastrophic forgetting on complex tasks, such as image classification.

\subsubsection{Continual Learning scenarios}
\label{subsec:CLscenarios}
In a real-world scenario, CL can be considered as the setting in which a model continuously receives input data and learn from each dataset until convergence. In a sequence of $T$ tasks, the model is required to learn from the batch of pairs ($X_t$, $Y_t$) from the $t^{th}$ task, $\forall t \in [0, ..., T]$. In this context, a \say{task} refers to a new training phase, distinct from previous ones, relying on new data with new input or output distributions. For instance, in multi-class classification, a new task may involve a batch of (input, output) pairs where all the outputs correspond to a class that the model has never encountered before. The difficulty of this incremental learning process can vary based on the distribution of the incremental batches and whether task identity is provided. 

The latter condition (i.e., whether the model has access to the task identity) is particularly relevant in CL: task ID's may or may not be available during training or testing, and may have to be inferred during testing. Depending on the scenario, this can make training more or less challenging. In the context of CL (or \textit{incremental learning}, IL) applied to computer vision, and particularly to the MNIST dataset, \cite{van2019three} and \cite{van2022three} outline three increasingly difficult scenarios: Task-IL, Domain-IL, Class-IL.

In \textit{Task-IL}, the easiest scenario, the model is given access to the ID of the task at hand. The single tasks have disjoint label spaces, meaning that one task includes samples from only one class. With task IDs provided, multi-headed models can be used, where each head is responsible for and trained on one task only. In this case, the real challenge is not to overcome catastrophic forgetting, but rather to extract meaningful shared feature representations across all the tasks. \textit{Domain-IL} is a scenario in which the input data distribution changes across tasks, but the label distribution is fixed. It is unnecessary to distinguish across tasks, because the outputs will always belong to the same set of classes regardless. Therefore, task ID's are not provided, nor does the model need to infer them at test time. For instance, this may be the case of an agent required to adapt to different environments. In this scenario, catastrophic forgetting can undermine the training, and it needs to be addressed with proper strategies. Finally, in \textit{Class-IL}, the model is required to learn how to distinguish a growing number of classes. Unlike in Task-IL, in Class-IL the task ID is not provided and has to be implicitly inferred at test time, requiring the model to solve both classification and task identification. Moreover, in this setting, the model must discriminate between classes never encountered in the same task (and that the model was never explicitly trained to compare). For these reasons, Class-IL is the most challenging scenario, reflecting real-world problems in which the agent needs to incrementally learn new classes of objects \citep{zhou2024class}.

\subsubsection{Continual Learning strategies}
The easiest way to overcome the limitations of catastrophic forgetting is to train the model from scratch on both new and old data, but this approach entails evident computational drawbacks, as well as privacy-related issues \citep{wang2024comprehensive}. In order to incrementally train models while mitigating catastrophic forgetting, several strategies have been proposed. A common technique is \textit{Replay} \citep{rolnick2019experience}, in which a subset of the previous training dataset is included in the new dataset. This allows the model to reinforce previous knowledge and adapt to the new and the old distribution. The selection of the training samples from the old task is critical: for example, \cite{rebuffi2017icarl} choose samples that are closest to the dataset average feature distribution, ensuring that they are representative of the whole dataset. Other strategies penalize changes to weights that result to be critical for previously learned tasks, thereby reducing the likelihood of forgetting important information.

Despite the relevance of these existing strategies, our research investigates CL settings in which architectures are not supported by any additional techniques. Our aim is to point out the main differences between MLP and KAN when these networks are trained in a CL scenario. For this reason, no CL strategies are employed in our study.

\subsection{State of the art on KAN in computer vision}
\cite{liu2024kan} introduce KANs and shows their applicability in several applications. However, the authors do not explore  how these architecture handle multi-dimensional inputs or computer vision tasks. More recent studies have addressed this gap. \cite{yu2024kan} carry out an comprehensive comparison between KANs and MLPs in several tasks and domains, from machine learning to natural language processing. In order to fairly compare the two types of models, the authors choose architectures that are comparable in terms of number of parameters and/or floating point operations. This choice ensures that both networks have equivalent computational power, although it is worth noting that MLPs and KANs use parameters in fundamentally different ways. The results from this comparison suggest that KANs outperform MLPs only in such tasks as symbolic formula regression, while they are generally weaker than MLPs in other areas,, including classic computer vision tasks such as image classification on the MNIST and CIFAR10 datasets. However, the equations used by \cite{yu2024kan} to compute the number of trainable parameters may contain errors; this is further discussed in Sec. \ref{subsec:architectures}. In another recent work, \cite{bodner2024ckan} investigate the performance of KAN-based convolutions compared to traditional convolutional networks in computer vision tasks. They find that KAN convolutions achieve acceptable accuracy levels on the MNIST dataset, while using up to seven times fewer parameters than traditional convolutional networks.

All these results are preliminary and further investigation is required to fully validate them, especially considering the still unknown role that hyper-parameters play in KANs. Regardless, these studies provide a solid foundation for exploring KANs' capabilities in continual learning settings within computer vision.



%% file: 3KANarch.tex
\section{Architectures and hyper-parameters}

\subsection{Choosing the right architectures for fair comparison}
\label{subsec:architectures}

Given the well-studied performance of MLPs on the MNIST dataset, we first defined the architecture of the MLP to be used in our study: [784, 784, 256, 10]. However, KANs differ fundamentally from MLPs, and no established guidelines currently exist to determine whether a KAN and an MLP model have comparable computational power. Key measures of computational ability include the number of learnable parameters and the number of floating point operations (FLOPs). In a recently published pre-print, \cite{yu2024kan} assess the number of trainable parameters in KANs as
\begin{equation}
\label{eq:paramKAN_w}
    params_{KAN} = \left(d_{in} \times d_{out}\right) \times (G+k+3) + d_{out},
\end{equation}
where $d_{in}$ and $d_{out}$ are the input and output dimensions of the single KAN layer, G is the grid size, and k is the degree of the single basis functions. The isolated $d_{out}$ addend represents the bias added to the outcomes of the single activation functions of a KAN layer ($\beta$ in Eq. \ref{eq:output_x}).


However, in the default PyKAN configuration, each connection needs $G+k+6$ trainable parameters, where $G+k$ is the number of coefficient that multiply each curve in the whole spline (\textit{coef} in code, or $c_i$ in Eq. \ref{eq:spline_x}). Of the remaining six parameters, two correspond to the scale factors for the trainable spline and the fixed activation function (\textit{scale\_base} and \textit{scale\_sp} in code, or $w_b$ and $w_s$ in Eq. \ref{eq:output_x}), while the other four parameters are associated with the symbolic KAN layer corresponding to each numerical KAN layer (\textit{sb\_batch} in code). This computation is line with the \textit{ptflops} library \citep{ptflops}, which calculates the theoretical amount of FLOPs in a neural networks, as well as the number of trainable parameters that the network needs. Given this, we conclude that Eq. \ref{eq:paramKAN_w} should be

\begin{equation}
\label{eq:paramKAN}
    params_{PyKAN} = \left(d_{in} \times d_{out}\right) \times (G+k+6) + d_{out},
\end{equation}
when the \textit{grid} and the \textit{mask} in PyKAN are not made trainable (as by the default PyKAN settings). 

EffKAN, on the other hand, is more lightweight than PyKAN. Both utilize two scaling weights for each connection (one for the trainable spline $w_s$ and one for the fixed activation function $w_b$) but, unlike PyKAN, EffKAN does not include a bias ($\beta$) or a symbolic layer, which adds 4 trainable parameters per connection in PyKAN. Therefore, the number of trainable parameters in an EffKAN layer can be thus computed:

\begin{equation}
\label{eq:paramKEffAN}
    params_{EffKAN} = \left(d_{in} \times d_{out}\right) \times (G+k+2).
\end{equation}

While this holds true in general, in order for KANs to work in CL scenarios, it is necessary to make $w_s$, $w_b$, and $\beta$ in Eq. \ref{eq:output_x} non-trainable, as explained in Sec. \ref{subsec:KANimplemCL} and Sec. \ref{subsec:EffKAN}. Additionally, the symbolic front in PyKAN can be disabled for efficiency purposes without affecting the optimization process. Because of this, the number of parameters to be considered when training PyKAN and EffKAN in CL settings is the same and is equal to:

\begin{equation}
\label{eq:paramKANCL}
    params_{KAN,CL} = \left(d_{in} \times d_{out}\right) \times (G+k).
\end{equation}

PyKAN and EffKAN with a shape of $[784, 128, 10]$, $G=5$, $k=3$, and no learnable scales or bias, have approximately  $813k$ parameters, which is comparable to the number of parameters in an MLP with a shape of $[784, 784, 256, 10]$ (around $818k$). Nevertheless, while fixing $w_s$, $w_b$, and $\beta$ is the best option for CL on 1D data, preliminary experiments on the MNIST dataset suggested that this choice may not be effective in higher-dimensional CL scenarios. In fact, neither EffKAN nor PyKAN was able to successfully perform this CL task. Because of this, \textbf{it was necessary to set $w_s$ as a trainable parameter}. This introduces one additional trainable parameter per connection in Eq. \ref{eq:paramKANCL} ($(G+k+1)$) both for PyKAN and EffKAN, bringing the total for the same model configuration ($[784, 128, 10]$, $G=5$, $k=3$) to about $914k$ trainable parameters. Therefore, \textbf{it was also necessary to scale up the MLP architecture} to $[784, 784, 285, 256, 10]$ (which only has about 300 parameters more than the chosen KAN architecture).

In terms of FLOPs, achieving a fair comparison is challenging. According to \cite{yu2024kan}, the described MLP requires about 1.63 million FLOPs, which is consistent with pftlop, whereas the described KAN has over 52.9 million FLOPs. This figure significantly exceeds the 101.6 kMAC reported by ptflop, which is probably due to the fact that some PyTorch operations used in PyKAN are unaccounted for by the library, which was designed for MLP, convolutional networks, and transformers. According to the formulation provided by \cite{yu2024kan}, to align the FLOPs of a KAN with those of the MLP (1.8M vs. 1.6M), its architecture should be $[28\cdot28,10]$ with grid size of 4 and spline order of 3. However, this architecture lacks hidden KAN layers, which limits deep data processing. Therefore, we avoid considering FLOPs for the moment, leaving this study for future work. Nevertheless, it is safe to state that PyKAN requires significantly more compute than EffKAN, even when the parameter $symbolic\_enabled$ is set to $False$. This adjustment boosts training speed in PyKAN, which nonetheless remains approximately 5-6 times slower than EffKAN.

Appendix \ref{app:ConvArchit} also contains a detailed analysis of traditional and KAN-based convolutional networks based on the number of trainable parameters.



\subsection{Training protocol: Class incremental learning and hyper-parameters}
\label{subsec:trainprot}

Each model was trained on the MNIST dataset according to the Class-IL protocol. The dataset is divided into five consecutive tasks based on labels: the first task includes classes [0, 1], the second task includes samples labeled as [2, 3], and so on. The models are trained on the samples from the subset $i$ only during the $i$-th task, after which the samples are drawn from the subset $i+1$. During training, the model is evaluated on a fixed test set that includes balanced random samples from all classes, averaging 1000 samples per class (with standard deviation of 59 across the ten classes). Consequently, at the end of the first task (training on images labeled as either 0 or 1), the maximum achievable accuracy on the test set it 20\%, since the model has no knowledge about 8 of the 10 classes in the test set. Ideally, a model that does not exhibit catastrophic forgetting should achieve $20 \times i$\% accuracy by the end of the $i$-th task.

An important factor to consider is the learning rate (LR). \cite{liu2024kan} optimized PyKAN on 1D regression tasks with LR set to 1. For training EffKAN on the MNIST dataset, LR is set to $10^{-3}$  \citep{efficient-kan}. Similarly, preliminary tests on MLP trained on MNIST suggested that $10^{-3}$ or $10^{-4}$ are optimal LRs. It’s worth noting that in these experiments, the models were not trained in a CL fashion and could be optimized on the whole dataset for many epochs. Therefore, to fully explore the power of PyKAN and EffKAN and compare them with MLP, we defined a training protocol where each architecture described in Sec. \ref{subsec:architectures} is trained several times with varying LRs ($10^{-3}$, $10^{-4}$, $10^{-5}$, $10^{-6}$). The protocol also included several LR decay factors ($0.6, 0.7, 0.8, 0.9, 1$), which multiply the LR at the beginning of each new task (with no scaling if the decay factor is set to 1). The choice of introducing a task-dependent decay factor is motivated by observations of model behavior during training. All the models (MLP and KANs) exhibited varying degrees of catastrophic forgetting. Reducing the LR across tasks can help balance plasticity and stability, allowing the model to retain knowledge from previous tasks while learning the current one. However, scaling the LR too aggressively may hinder learning in later tasks, particularly when the LR becomes as small as $10^{-7}$. Additionally, we consider another variable: the number of epochs per task. 

Overall, the training protocol for Class-IL involved a grid of four values for the LR ([$10^{-3}, 10^{-4}, 10^{-5}, 10^{-6}$]), five decay factor ([$0.6, 0.7, 0.8, 0.9, 1$]), and ten epoch values ([$1, 2, 3, 4, 5, 6, 7, 8, 9, 10$]). In light of this hyper-parameter grid, each architecture was trained 200 times. 


%% file: 4results.tex
\section{Results}
\label{sec:results}

\begin{table}[ht]
    \centering
    \begin{tabular}{c|c||c|c|c||c}
        Model   & Architecture              & Epochs    & LR        & LR decay  & Max Acc \\ \hline
        MLP     & [784, 784, 285, 256, 10]  & 2         & $10^{-5}$ & 0.7       & 0.40 \\
        PyKAN   & [784, 128, 10], G=5, k=3  & 7         & $10^{-4}$ & 0.6       & 0.28 \\
        EffKAN  & [784, 128, 10], G=5, k=3  & 5         & $10^{-5}$ & 0.8       & 0.52\\
    \end{tabular}
    \caption{Optimal hyper-parameter configuration for each model, with the corresponding highest test accuracy value during the fifth and last task.}
    \label{tab:bestacc}
\end{table}

We trained the three baseline architectures discussed in Sec. \ref{subsec:architectures}: an MLP, a PyKAN, and an EffKAN. Since the experimental protocol described in Sec. \ref{subsec:trainprot} includes 200 training runs per model, reporting all the obtained results would be impractical and counterproductive. Therefore, we only present the highest-performing training for each model, regardless of whether the hyper-parameters (number of per-task-epochs, LR, and LR decay factor) are consistent across the models. This comparison does not lack fairness as every possible combination of these three hyper-parameters was evaluated for each model. Performance is assessed based on the highest test accuracy during the last task, the fifth one, when the model has had the opportunity to learn all classes in the dataset. As explained in Sec. \ref{subsec:trainprot}, an ideal model should exhibit a 20\% increase in accuracy with each new task, eventually reaching 100\%. However, Class-IL is a challenging scenario and such ideal behavior is not expected. Table \ref{tab:bestacc} shows the optimal configuration for each network and the corresponding best accuracy value on the test set. 

\begin{figure}[ht]
    \centering
    \begin{minipage}[b]{0.72\textwidth}
        \centering
        \includegraphics[width=\linewidth]{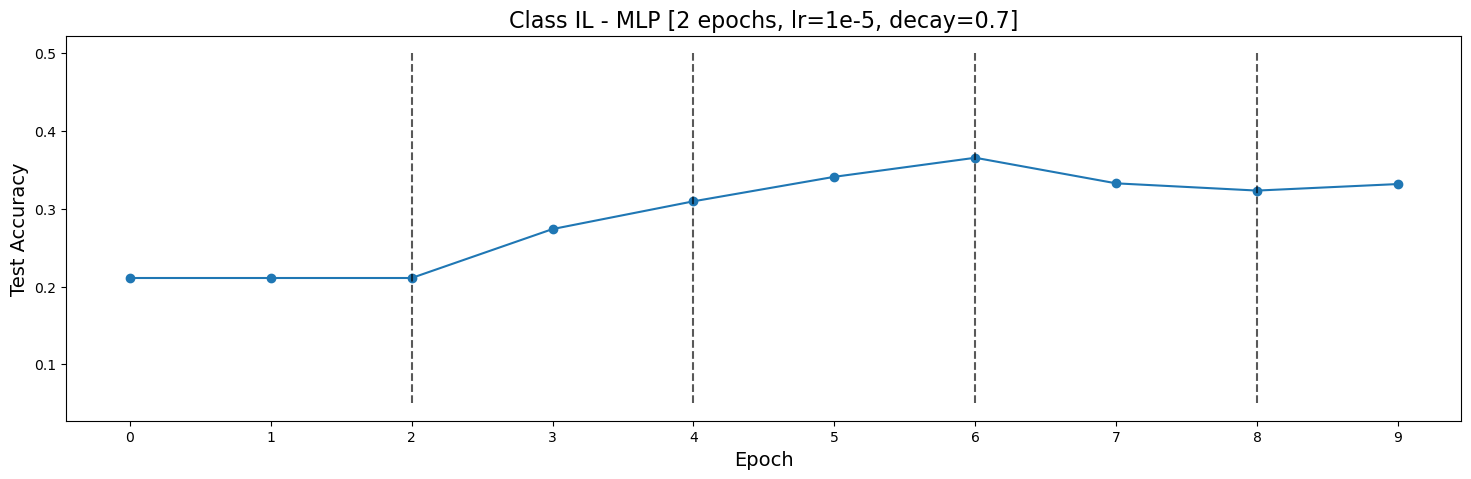}
        \subcaption{Test accuracy plot for MLP.}\label{fig:accplotMLP}
    \end{minipage}
    \begin{minipage}[b]{0.27\textwidth}
        \centering
        \includegraphics[width=\linewidth, trim=1cm 1cm 1.8cm 1.5cm, clip]{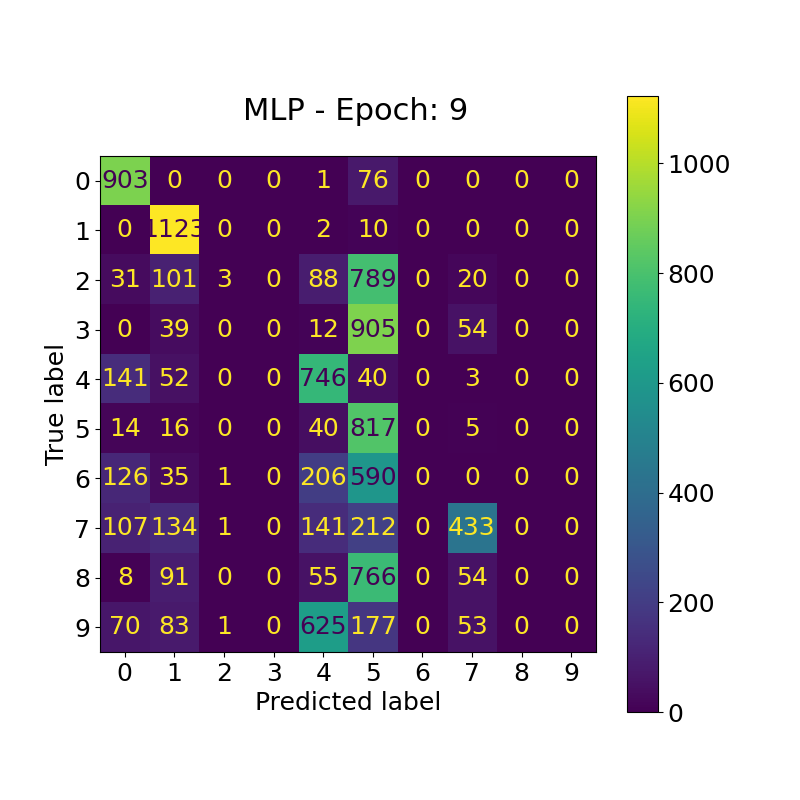}
        \subcaption{Confusion matrix for MLP.}\label{fig:confmatrMLP}
    \end{minipage}
    
    \begin{minipage}[b]{0.72\textwidth}
        \centering
        \includegraphics[width=\linewidth]{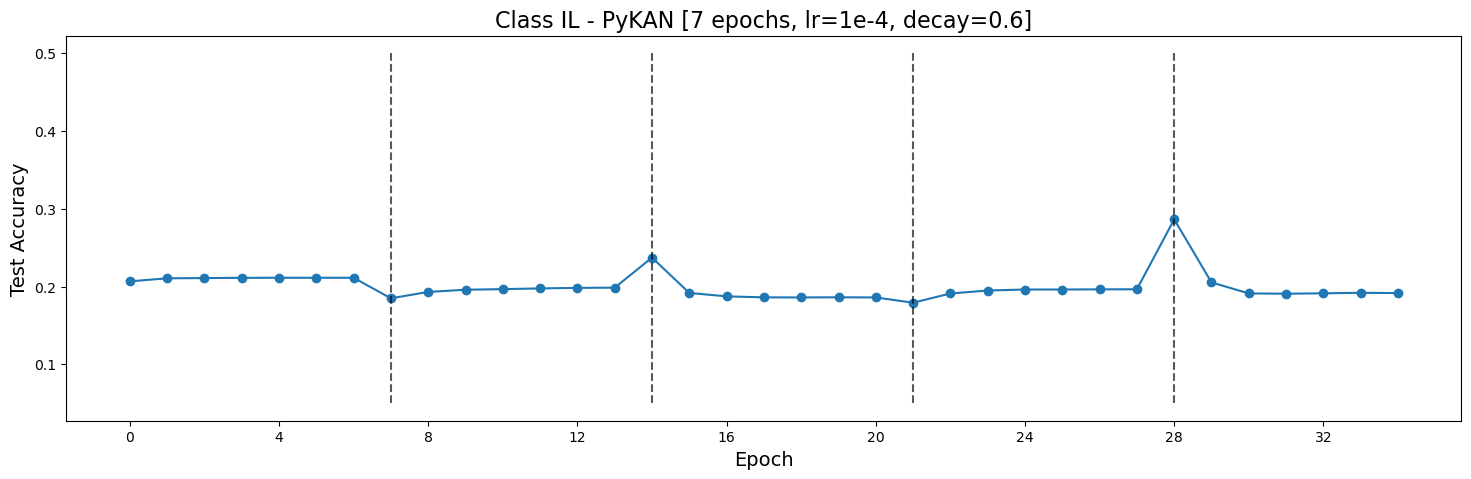}
        \subcaption{Test accuracy plot for PyKAN.}\label{fig:accplotPyKAN}
    \end{minipage}
    \begin{minipage}[b]{0.27\textwidth}
        \centering
        \includegraphics[width=\linewidth, trim=1cm 1cm 1.8cm 1.5cm, clip]{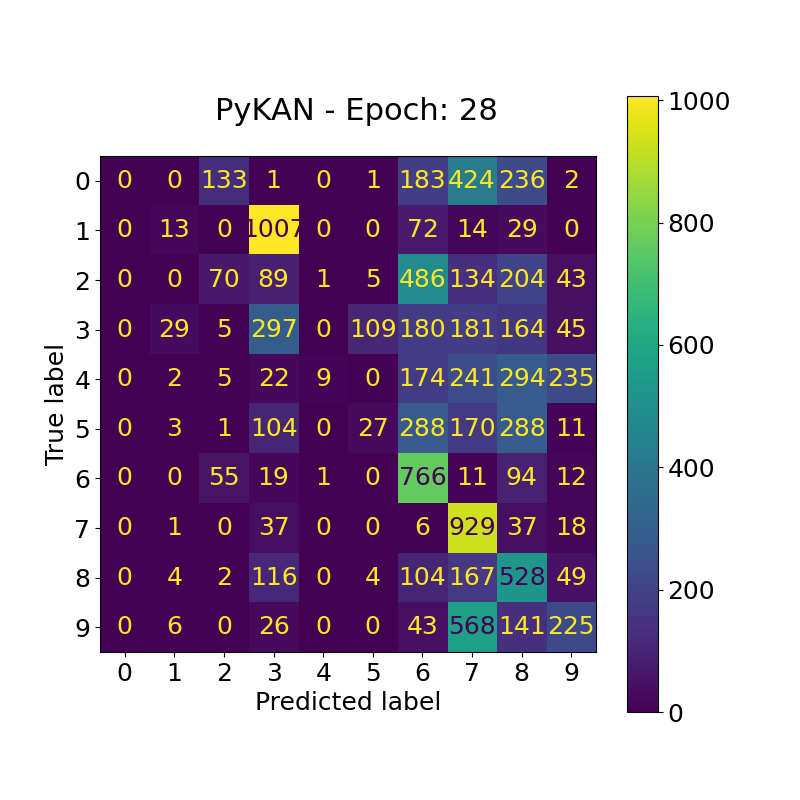}
        \subcaption{Confusion matrix for PyKAN.}\label{fig:confmatrPyKAN}
    \end{minipage}
    
    \begin{minipage}[b]{0.72\textwidth}
        \centering
        \includegraphics[width=\linewidth]{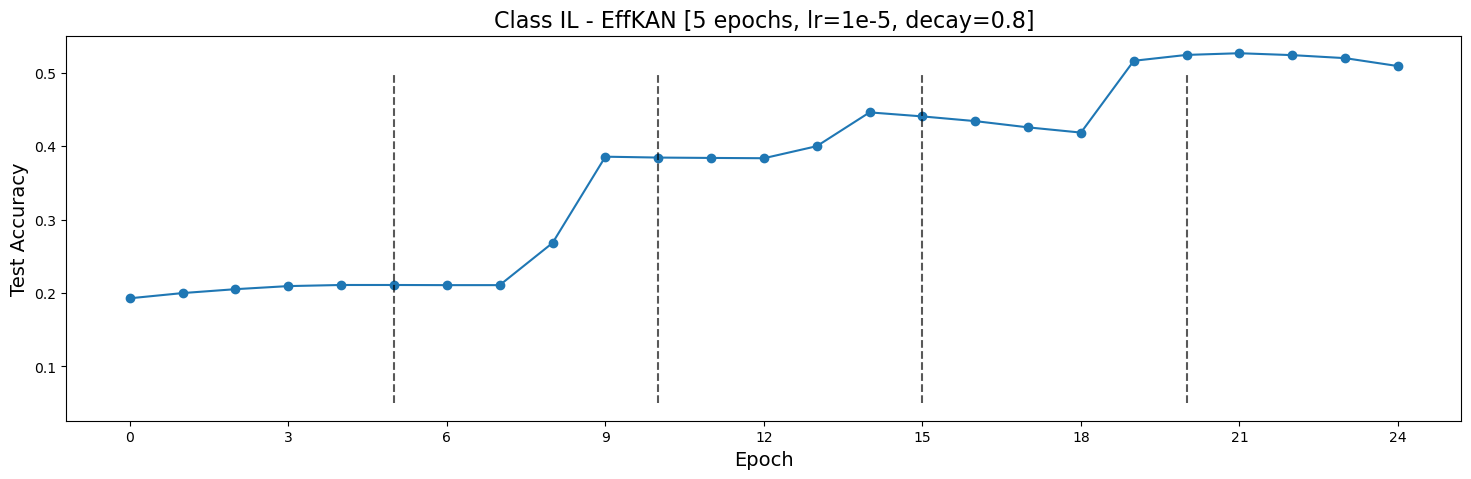}
        \subcaption{Test accuracy plot for EffKAN.}\label{fig:accplotEffKAN}
    \end{minipage}
    \begin{minipage}[b]{0.27\textwidth}
        \centering
        \includegraphics[width=\linewidth, trim=1cm 1cm 1.8cm 1.5cm, clip]{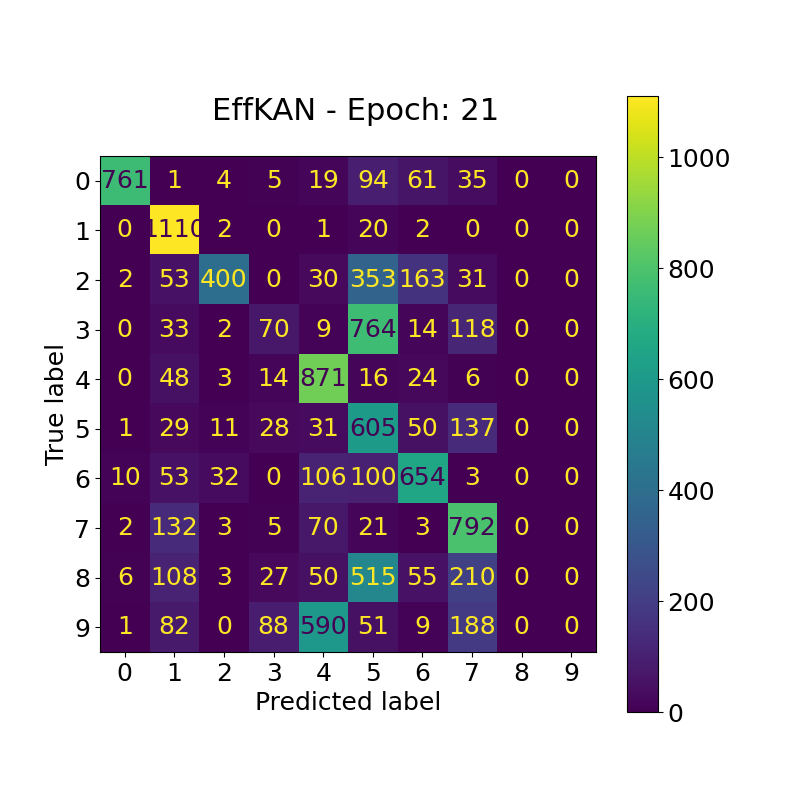}
        \subcaption{Confusion matrix for EffKAN.}\label{fig:confmatrEffKAN}
    \end{minipage}
    
    \caption{Test accuracy curves and confusion matrices for the three considered models, each in its optimal hyper-parameter configuration. The dashed vertical lines in the plots represent the beginning of a new task. The confusion matrix represent the test performance at the epoch where the corresponding model shows highest test accuracy in the final task.}
    \label{fig:plots&matrices}
\end{figure}



Among 200 hyper-parameter configurations, EffKAN emerged as the highest-performing model, achieving 52\% accuracy on the test set. MLP is the second-best model, with 40\% test accuracy, while PyKAN recorded the lowest at 28\%. In order to fully comprehend and discuss these results, Fig. \ref{fig:plots&matrices} presents the test accuracy curves for the three models, each in its optimal hyper-parameter configuration (figures \ref{fig:accplotMLP}, \ref{fig:accplotPyKAN}, and \ref{fig:accplotEffKAN}). The dashed lines in these plots represent the epoch at which a new task begins, determined by the number of epochs per task. Moreover, Fig. \ref{fig:plots&matrices} includes the confusion matrices representing the maximum test accuracy for each model (epoch 9 for MLP, Fig. \ref{fig:confmatrMLP}; epoch 28 for PyKAN, Fig.\ref{fig:confmatrPyKAN}; epoch 21 for EffKAN, Fig. \ref{fig:confmatrEffKAN}).

The test accuracy plots are particularly useful in order to evaluate whether a model is, in fact, learning from new tasks. In the next Section, we will analyze the behavior of each model on an epoch-by-epoch basis using confusion matrices. From the comparison between the three plots (Figures \ref{fig:accplotMLP}, \ref{fig:accplotEffKAN}, and \ref{fig:accplotPyKAN}), EffKAN displays the most \textit{ideal} behavior, with accuracy consistently increasing as new tasks are introduced. However, the positive slope is only constant in the fist two tasks (+20\%), after which it decreases: +5\% during task 3, +7\% during task 4, until it reaches 0 in the last task. Regardless, the final accuracy value (52\%) is considerably far from a random prediction (10\% on a 10-class dataset), albeit not comparable to state-of-the-art performance on the MNIST dataset.

Appendix \ref{app:ConvArchit} reports the results from the preliminary study on traditional neural networks compared to KAN-based convolutional neural networks.



%% file: 5discussion.tex
\section{Discussions}
\label{sec:discussions}

\begin{figure}
    \centering
    \centering
    \begin{minipage}[b]{0.30\textwidth}
        \centering
        \includegraphics[width=\linewidth, trim=1cm 1cm 1.8cm 1.5cm, clip]{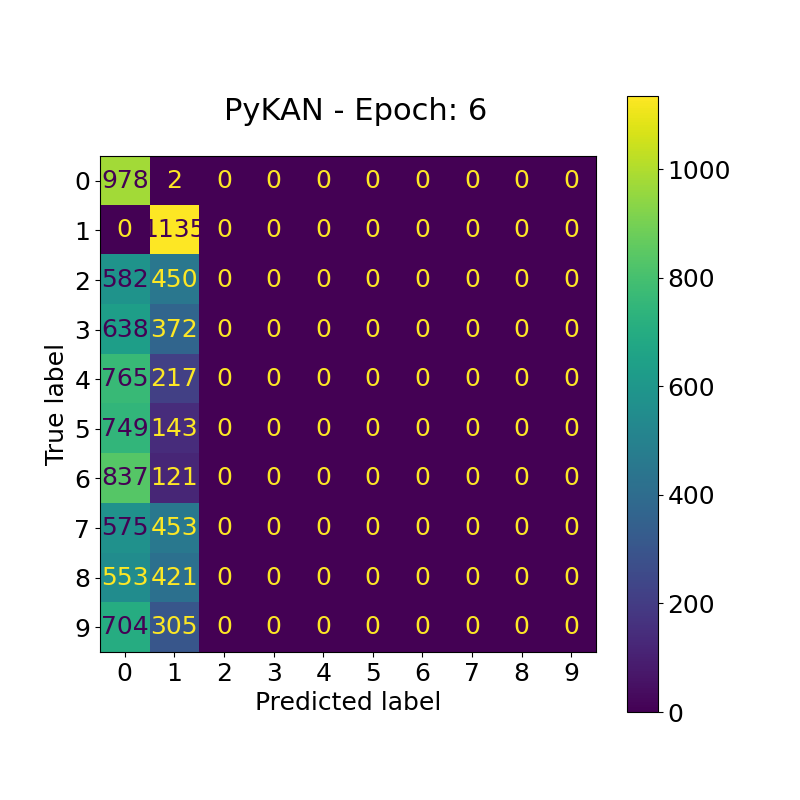}
        \subcaption{End of task 1.}\label{fig:confmatrPyKAN_ep6}
    \end{minipage}
    \begin{minipage}[b]{0.30\textwidth}
        \centering
        \includegraphics[width=\linewidth, trim=1cm 1cm 1.8cm 1.5cm, clip]{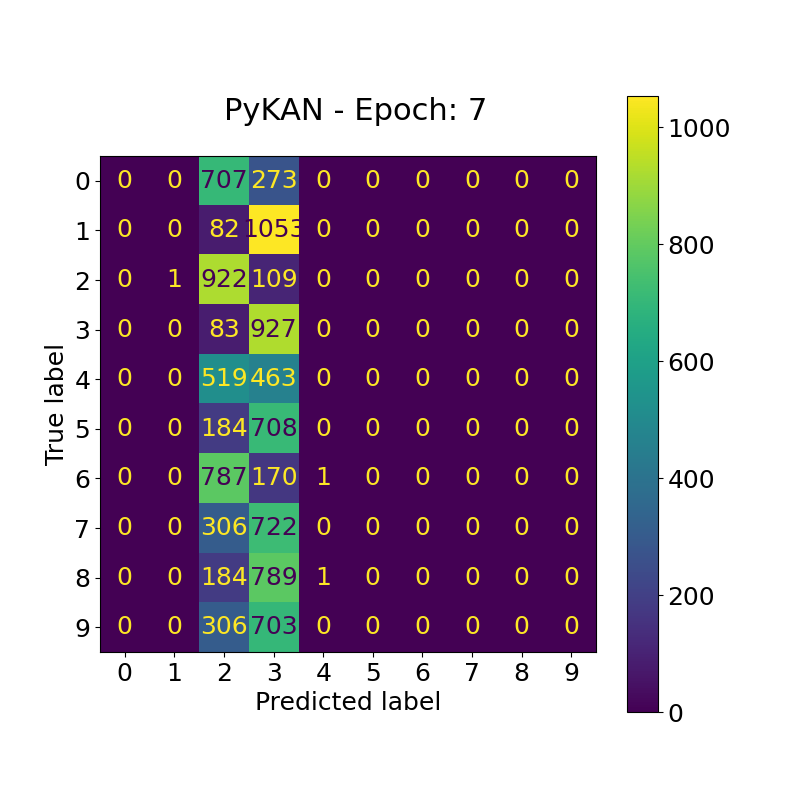}
        \subcaption{Beginning of task 2.}\label{fig:confmatrPyKAN_ep7}
    \end{minipage}
    \begin{minipage}[b]{0.30\textwidth}
        \centering
        \includegraphics[width=\linewidth, trim=1cm 1cm 1.8cm 1.5cm, clip]{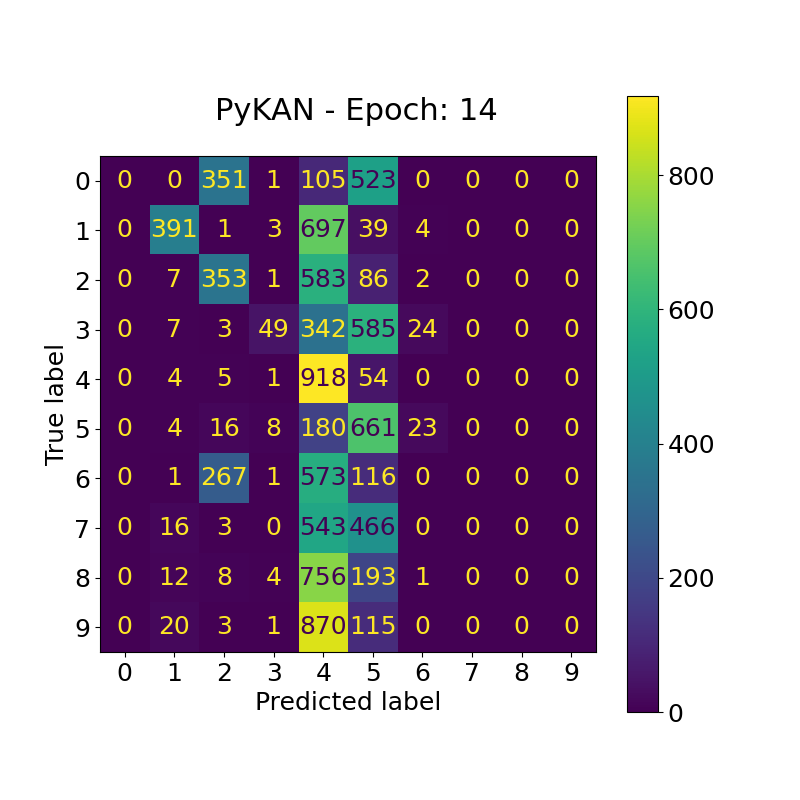}
        \subcaption{Beginning of task 3.}\label{fig:confmatrPyKAN_ep14}
    \end{minipage}
    \caption{Confusion matrix for PyKAN at the end of task 1 \ref{fig:confmatrPyKAN_ep6}, at the beginning of task 2 \ref{fig:confmatrPyKAN_ep7}, and at the beginning of task 3 \ref{fig:confmatrPyKAN_ep14}.}
    \label{fig:confmatrPyKAN_task1to2}
\end{figure}

The test accuracy plots show the epoch-wise trend in the knowledge learnt by each model. However, these plots may be misleading because of the stability-plasticity dilemma. For example, it would be misleading to conclude that PyKAN (Fig. \ref{fig:accplotPyKAN}) fails to learn after task 1. A decrease in accuracy in CL scenarios typically indicates that the model is forgetting more than it is learning. The plot in Fig. \ref{fig:accplotPyKAN} shows that PyKAN, like the other models, achieves perfect performance during task 1, reaching 20\% accuracy—the highest possible, given that the model has only seen 2 out of 10 classes. Moving along the x-axis to task 2, PyKAN's performance slightly drops but remains around 20\% at epoch 7. A clearer picture of what is happening in this moment can be obtained from the per-epoch confusion matrices\footnote{The whole set of confusion matrices for the best-scoring training runs can be found in out GitHub repository.} in Fig. \ref{fig:confmatrPyKAN_task1to2}. At epoch 7 (beginning of task 2, Fig. \ref{fig:confmatrPyKAN_ep7}), PyKAN appears to completely overwrite the knowledge gained during task 1 (epoch 6, Fig. \ref{fig:confmatrPyKAN_ep6}). Similarly, at the onset of task 3 (Fig. \ref{fig:confmatrPyKAN_ep14}), samples labeled as "4" and "5" prevail in the confusion matrix. The model seems to readily forget most knowledge about the classes "2" and "3" and to restore some knowledge about class "1", which it later forgets during task 3. This behavior shows in Fig. \ref{fig:confmatrPyKAN}, where PyKAN achieves its best performance. At the beginning of the last task, the model restores knowledge about previous classes and shows a higher degree of sparsity in the confusion matrix compared to earlier epochs, though not along the diagonal.

\begin{figure}
    \centering
    \centering
    \begin{minipage}[b]{0.30\textwidth}
        \centering
        \includegraphics[width=\linewidth, trim=1cm 1cm 1.8cm 1.5cm, clip]{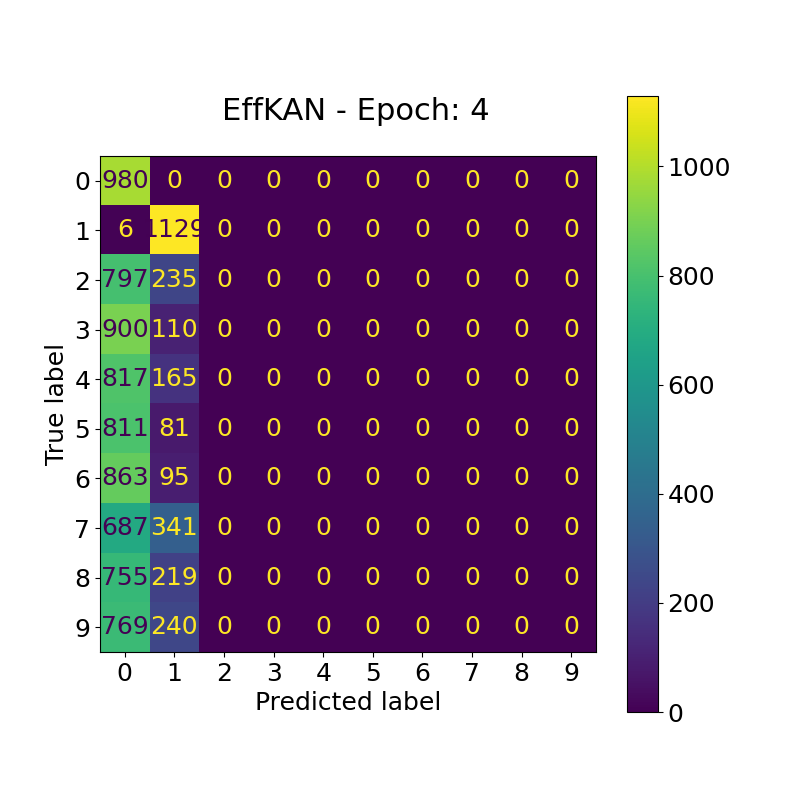}
        \subcaption{End of task 1.}\label{fig:confmatrEffKAN_ep4}
    \end{minipage}
    \begin{minipage}[b]{0.30\textwidth}
        \centering
        \includegraphics[width=\linewidth, trim=1cm 1cm 1.8cm 1.5cm, clip]{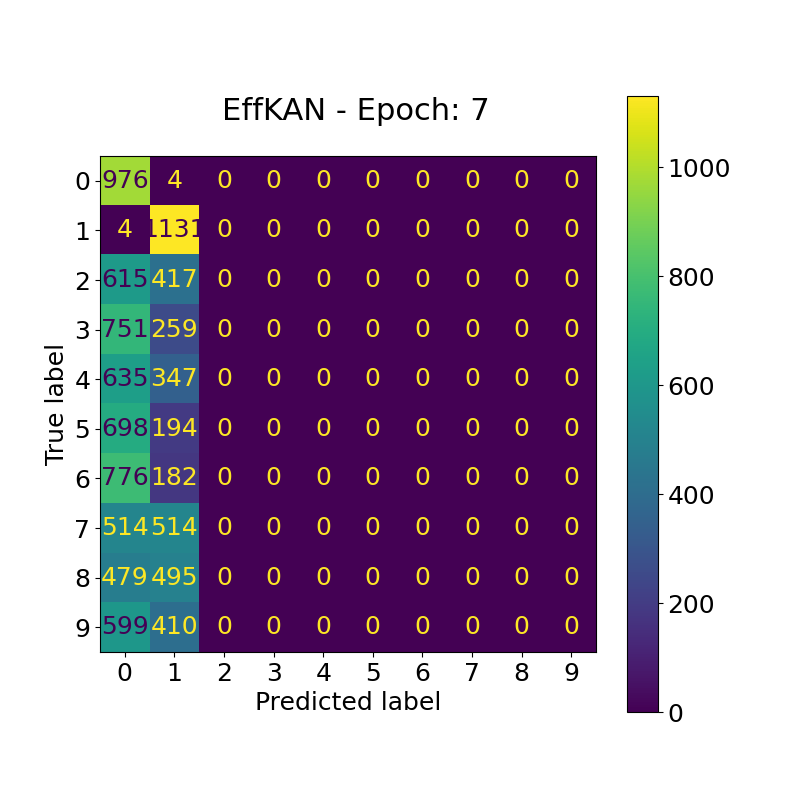}
        \subcaption{Third epoch of task 2.}\label{fig:confmatrEffKAN_ep7}
    \end{minipage}
    \begin{minipage}[b]{0.30\textwidth}
        \centering
        \includegraphics[width=\linewidth, trim=1cm 1cm 1.8cm 1.5cm, clip]{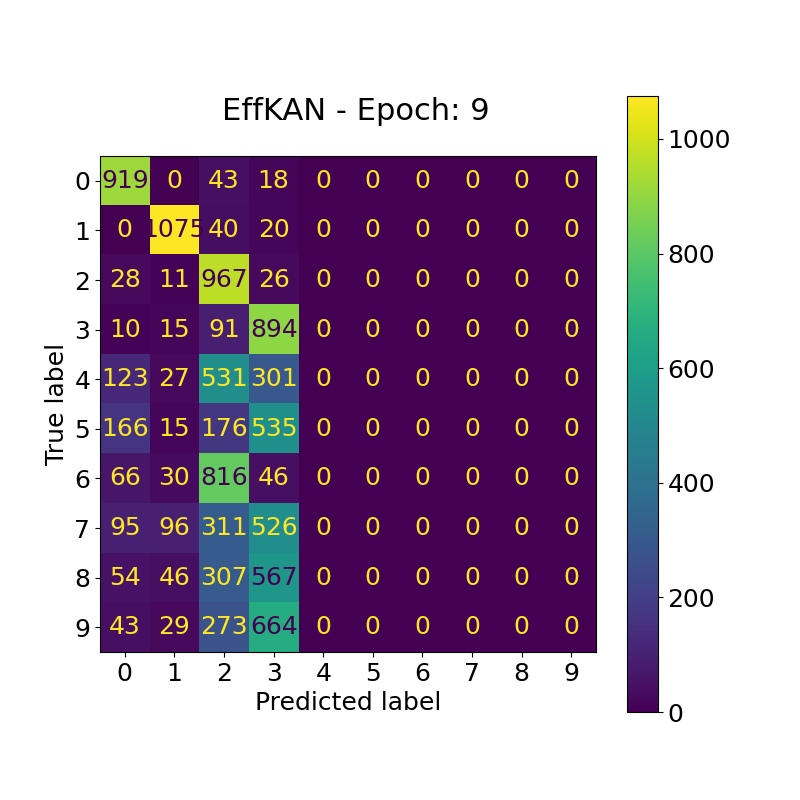}
        \subcaption{End of task 2.}\label{fig:confmatrEffKAN_ep9}
    \end{minipage}
    \caption{Confusion matrix for EffKAN at the end of task 1 \ref{fig:confmatrPyKAN_ep6}, at the third epoch of task 2 \ref{fig:confmatrPyKAN_ep7}, and at the end of task 2 \ref{fig:confmatrPyKAN_ep14}.}
    \label{fig:confmatrEffKAN_task1to2}
\end{figure}

Similarly, we can analyze the behavior of EffKAN. From the plot in Fig. \ref{fig:accplotEffKAN}, EffKAN seems to have learnt as much as possible from tasks 1 and 2 (40\% accuracy), but it is interesting to observe how slowly it learns. Fig. \ref{fig:confmatrEffKAN_task1to2} shows the confusion matrices for EffKAN at the end of task 1 (Fig. \ref{fig:confmatrEffKAN_ep4}), at the third epoch of task 2 (Fig. \ref{fig:confmatrEffKAN_ep7}), and at the end of task 2 (Fig. \ref{fig:confmatrEffKAN_ep9}). After three epochs in task 2, EffKAN maintains its knowledge from the first task while failing to acquire new knowledge from the second task, indicating a high degree of stability. However, by the end of task 2 (epoch 9), the model has mastered both tasks. As the model transitions to the third task, it starts exhibiting signs of forgetting, losing significant knowledge about the second task. Consequently, accuracy increases by only 5\% from epoch 12 to 14 and slightly decreases from epoch 14 to 15. While the model's stability remains consistent—requiring several epochs to learn new tasks—its plasticity diminishes as more tasks are introduced. The extent of knowledge loss seems to increase with each successive task. Eventually, as can be seen in Fig. \ref{fig:confmatrEffKAN}, EffKAN does not learn anything about the last task (as the last two columns are completely empty), and starts predicting "8" or "9" only at the very last epoch. This may be also attributed to the LR decay, because of which the LR is reduced to $4 \times 10^{-6}$ in the final task. 

MLP exhibits more erratic behavior than EffKAN, akin to that of PyKAN. MLP does not seem to have any stability, just like PyKAN, and it tends to lose knowledge from previous tasks as soon as a new one begins. It is worth noting that the confusion matrix at its final and best epoch (Fig. \ref{fig:confmatrMLP}) shows that a good amount of samples are incorrectly classified as "1" and "2", while the model seems unable to predict labels from the second task. This may be due to the optimal number of epochs per task being 2, which is relatively short for computer vision tasks, and the model may not have enough time to redistribute its predictions across classes. Despite this, we cannot discard the hypothesis that MLP may be more biased towards the first task than KAN architectures. Additionally, MLP appears less consistent than EffKAN and PyKAN, being less likely to reach the same (or similar) parameter configuration across different runs, thus making experiments less repeatable. This claim, however, is based on the observation of few examples, and deeper investigation would be necessary to corroborate it.

All models have around $914k$ parameters. While training time for MLP and EffKAN is comparable, taking around 1-2 seconds per epoch, training PyKAN requires significantly longer, around 16 seconds per epoch on the same hardware. It is important to note that the original PyKAN framework has strict requirements for training and testing datasets, allowing input only in the form of Python dictionaries. This necessitates that the entire dataset be loaded into memory, which can be impractical or unfeasible for larger datasets like MNIST. In order to overcome this limitation, we trained PyKAN with the same custom training loop used for MLP and EffKAN. However, this approach meant omitting the regularization loss present in the original PyKAN framework, which may have adversely affected its performance. Interestingly, in contrast to 1D data tasks, both EffKAN and PyKAN struggled to learn effectively in this Class-IL scenario unless scale weights were included in the optimization process ($w_s$ in Eq. \ref{eq:output_x}). Also, the optimal hyper-parameter configuration for all the three models featured a LR decay factor < 1. The best decay factor (0.6) and LR ($1e-4$) for PyKAN are particularly low considering that, in the original framework, the LR is set to $1.0$ for 1D regression tasks. 

One of the greatest strengths of KANs is their locality, which enables them to disentangle subsets of the input domain and to learn differently for each subset using locally-defined spline functions. This property is particularly advantageous for 1D input data. However, in higher-dimensional inputs, the benefits of locality may diminish. While individual pixels can be viewed as 1D input, they are not independent from one another. Computer vision relies on the relative positions of pixels and their combined values to carry out any visual task, like image classification or object detection. Hence, the interconnection of multi-dimensional inputs, like pixels in images, challenges the efficacy of locality in KANs.

%% file: 6conclusions.tex
\section{Conclusions}

Kolmogorov-Arnold Networks are a novel class of trainable neural models emerging from the deep learning research community. This study is intended as a preliminary investigation into the behavior of KAN-based models in computer vision CL scenarios. We conducted a fair comparison between KAN and MLP architectures, designed to have the same number of trainable parameters. Our results, supported by an extensive hyper-parameters grid search, hint that KAN models might be a good alternative to MLP, and the theoretical foundation of KANs, based on the relaxed Kolmogorov-Arnold theorem, appears robust for real-world applications. EfficientKAN, in particular, outperformed MLP in the proposed class-IL tasks, demonstrating greater versatility compared to the original PyKAN framework, which, nonetheless, demands substantially more computational resources and longer training times. This may be due to the fact that PyKAN is a framework designed to train KANs that can be pruned and interpreted (i.e., expressed via a symbolic representation), and much of the computational demand is absorbed by the operations required to compute and store pre- and post-activations. We did not explore the possibility to prune or find the symbolic representation of the trained KAN models, because these aspects are not relevant in the field of CL and because EfficientKAN lacks these features. Moreover, with $1e5$ activation functions, the symbolic representation of the trained PyKAN would be meaningless.

In conclusion, our findings indicate that KANs are a promising alternative to MLPs in this specific context. Further investigation is necessary to fully evaluate this new kind of neural network. We hope that our work can support the research community in this journey.

%% file: 9appendix.tex
\section{Assessing the equivalence between PyKAN and EfficientKAN}
\label{app:PyvsEff}

\begin{figure}[!ht]
    \centering
    \includegraphics[width=0.9\linewidth]{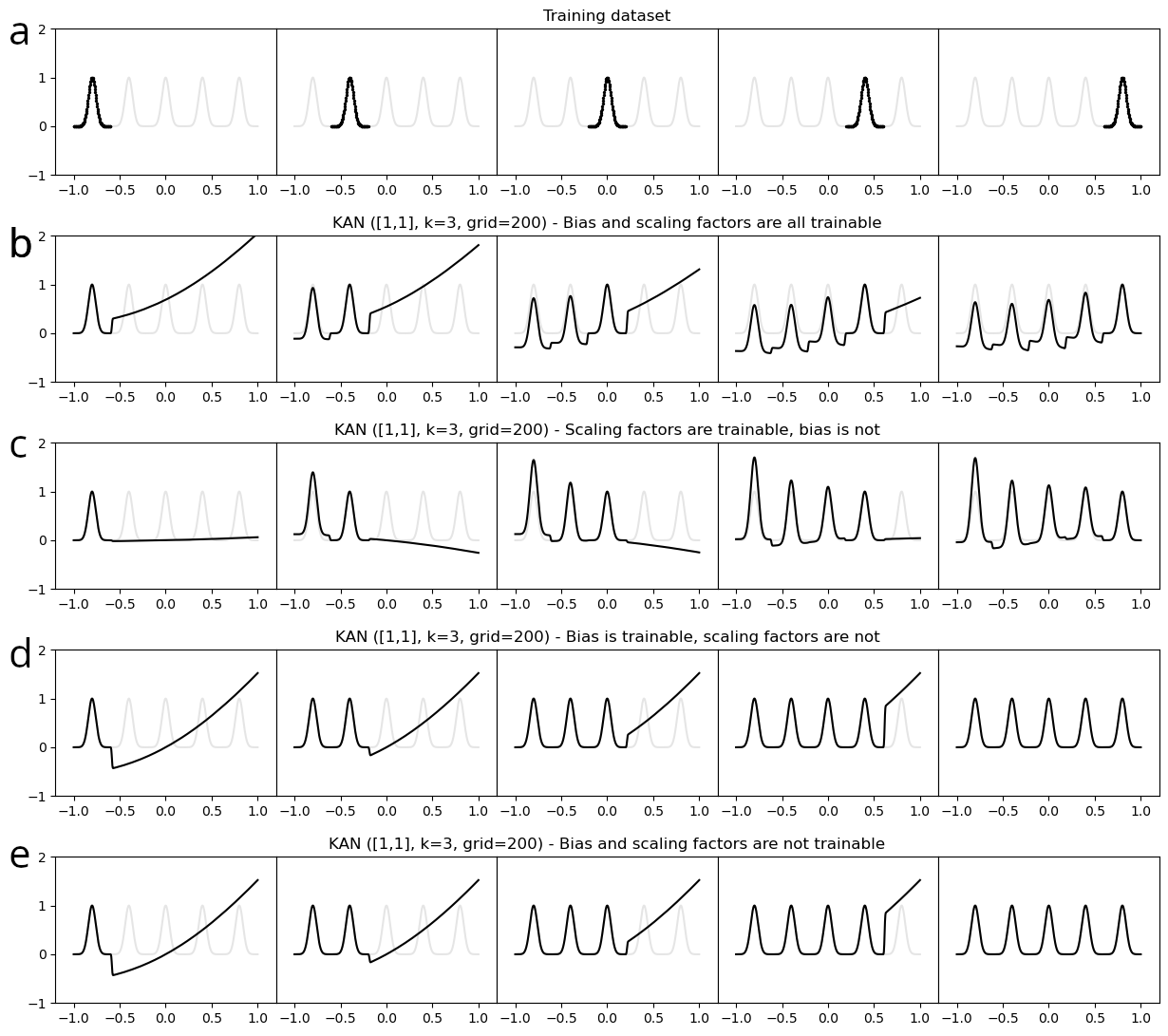}
    \caption{Different configurations of continual learning on a toy dataset containing five consecutive Gaussian peaks, as can be seen in the plot \textit{a}. The points are sequentially presented to a [1,1] PyKAN model with grid size set to 200 and spline order to 3. As can be seen from the plots \textit{b} and \textit{c}, optimizing the scale factors is detrimental for toy continual learning scenarios, whereas the choice of whether or not to optimize the bias factor might be irrelevant (\textit{d} vs \textit{e}).}
    \label{fig:CL_toy_dataset_trainable}
\end{figure}

\begin{figure}[!ht]
    \centering
    \includegraphics[width=0.9\linewidth]{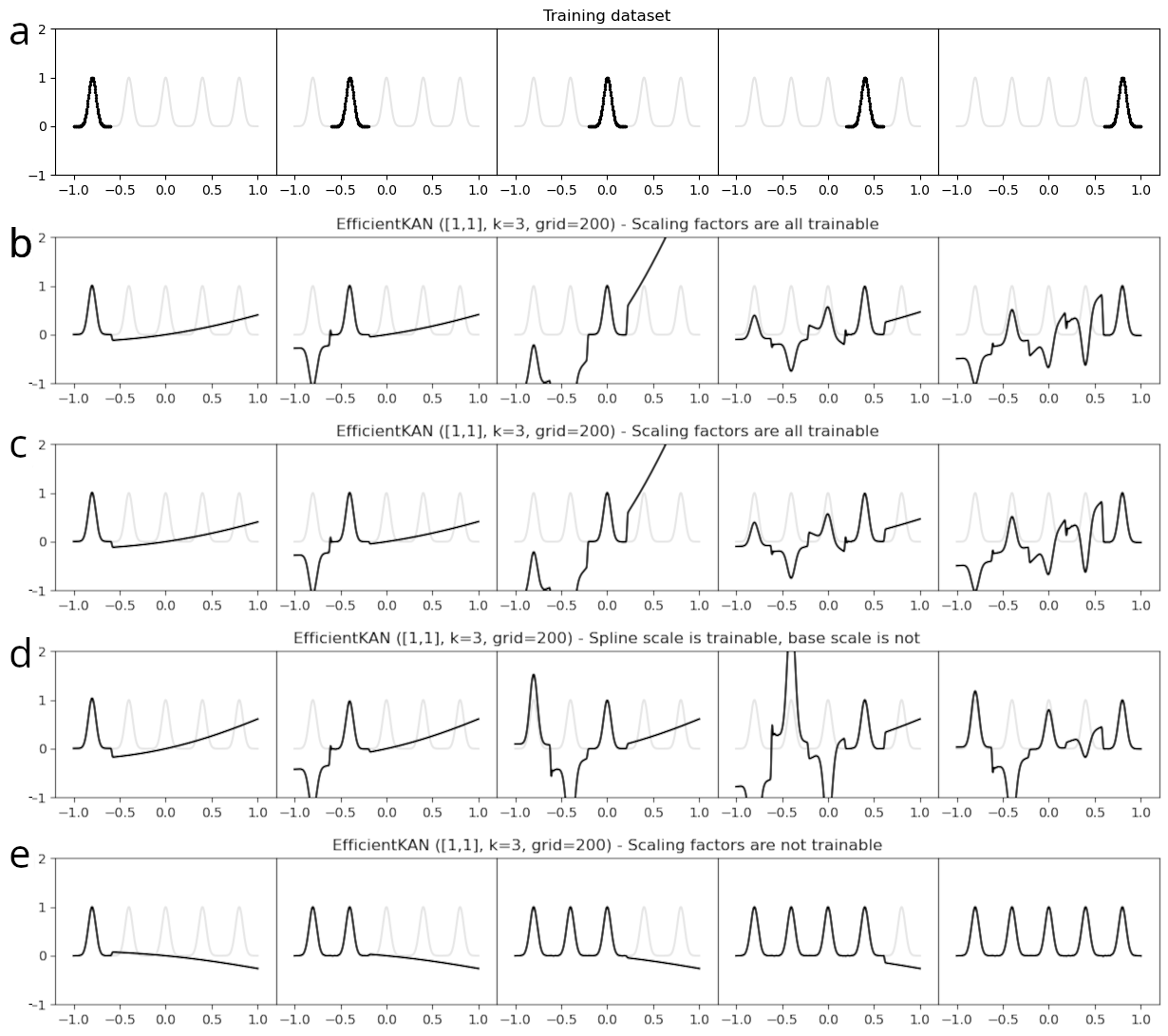}
    \caption{EfficientKAN trained and tested on the same toy dataset described in Sec. \ref{subsec:KANimplemCL}. The network has the same architecture, grid size, and spline order as the KAN used in Fig. \ref{fig:CL_toy_dataset_trainable} ([1,1], 200, 3) and is trained by setting the scale factors as non-trainable parameters. Unlike PyKAN, however, activation functions in EfficientKAN do not include bias factors, which were also set as non-trainable parameters for the toy experiment in Fig. \ref{fig:CL_toy_dataset_trainable}.}
    \label{fig:CL_toy_dataset_EffKAN}
\end{figure}

The plots in Fig. \ref{fig:CL_toy_dataset_trainable} show that optimizing the scale factors strongly hinders the ability of PyKAN to fit data distributions from sequential tasks. In particular, if scale factors are optimized (Fig. \ref{fig:CL_toy_dataset_trainable}b and c), the training is not locally \say{locked}, meaning that training on new data points affects distant parameters. More precisely, if the bias term is also trainable, the model retains the shape of previously regressed functions but shifts them along the y-axis (Fig. \ref{fig:CL_toy_dataset_trainable}b). In contrast, if the bias term is fixed, each new task changes the previously-regressed functions in terms of shape (Fig. \ref{fig:CL_toy_dataset_trainable}c), with the first peak reaches higher values as more tasks are introduced. However, when the scale factors are fixed, the KAN correctly fits the training data points without forgetting previous knowledge, regardless of whether the bias term is optimizable (Fig. \ref{fig:CL_toy_dataset_trainable}d) or not (Fig. \ref{fig:CL_toy_dataset_trainable}e). It should be noted that the limited contribution of the bias in this setting is probably due to the fact that the model has one edge only, i.e., one trainable activation function.

Fig. \ref{fig:CL_toy_dataset_EffKAN} shows the results from the same toy experiment described in Sec. \ref{subsec:KANimplemCL} (five consecutive Gaussian peaks shown one by one to the model) applied to an EfficientKAN [1,1] with grid size set to 200 and spline order to 3. By fixing the scale factors $w_b$ and $w_s$, EfficientKAN can accurately model the training data points without displaying catastrophic forgetting-related behaviors (Fig. \ref{fig:CL_toy_dataset_EffKAN}a). Similarly to PyKAN, if $w_b$ and $w_s$ are involved in the optimization process (Fig. \ref{fig:CL_toy_dataset_EffKAN}b-d), EfficientKAN struggles in this simple CL experiment. 

This comparison proves that PyKAN and EfficientKAN work equally well only in easy 1D regression tasks within CL scenarios. Further investigation is needed to fully examine the behavior of KANs (in either implementation) for multi-dimensional datasets.

\section{KAN-based convolutional networks}
\label{app:ConvArchit}

The Convolutional KANs (CKAN) repository \citep{ckan} introduces KAN-based convolutions, theorized by \cite{bodner2024ckan}, which differ from traditional convolutions in that each entry of the input matrix is activated by a learnable non-linear function, instead of being multiplied by learnable scalar values. In this sense, KAN kernels can be considered as KAN linear [4, 1] layers, receiving four inputs and producing one output value. 

Let an $m \times n$ input image be defined as
\begin{equation}
    img = 
    \begin{bmatrix}
        p_{11} & p_{12} & \hdots & p_{1n} \\
        p_{21} & p_{22} & \hdots & p_{2n} \\
        \vdots & \vdots & \ddots & \vdots\\
        p_{m1} & p_{m2} & \hdots & p_{mn} \\
    \end{bmatrix}
\end{equation}

and a $2 \times 2$ KAN-based convolutional kernel as
\begin{equation}
    \Phi = 
\begin{bmatrix}
    \phi_{11}(\cdot) & \phi_{12}(\cdot) \\
    \phi_{21}(\cdot) & \phi_{22}(\cdot)
\end{bmatrix}
.
\end{equation}

For each input \( x_i \), a learnable spline-based function \( \phi_i \) is applied, and the resulting value from the convolution step is \(\Phi(x)\). 

For classification tasks, several configurations of KAN-based convolutional neural networks can be designed. 

\subsection{Architectural choices for KAN-convolutional networks}
\label{subapp:archKANV}

As explained in Sec. \ref{subsec:architectures}, a linear KAN (or EffKAN) layer requires
\begin{equation}
\label{eq:paramKANCL_app}
    params_{KAN,CL} = \left(d_{in} \times d_{out}\right) \times (G+k)
\end{equation}
trainable parameters when every bias and scale factor is fixed to the initialization value. An MLP linear layer would need
\begin{equation}
\label{eq:paramMLP_app}
    params_{MLP} = \left(d_{in} \times d_{out}\right) + d_{out}
\end{equation}
trainable parameters.

As for convolutional layers, a traditional convolution needs 
\begin{equation}
\label{eq:paramConv_app}
    params_{conv} = n_f \times \left(k_s^2 + 1\right)
\end{equation}
trainable parameters (where $n_f$ is the number of filters in the kernel and $k_s$ is the kernel size) while a KAN-based convolutional layer uses
\begin{equation}
\label{eq:paramKConvFix_app}
    params_{CKAN} = n_f \times k_s^2 \times \left(G + k + 2\right)
\end{equation}
trainable parameters if bias and scales are optimizable, and
\begin{equation}
\label{eq:paramKConvTrain_app}
    params_{CKAN,CL} = n_f \times k_s^2 \times \left(G + k\right)
\end{equation}
otherwise.

Linear KAN and EfficientKAN layers have already been studied in the context of CL, with results indicating that fixing parameters such as bias and scale factors enhances these networks' ability to learn continuously. While this holds true for KAN, a similar investigation should be conducted to assess whether making these parameters opimizable is detrimental for KAN convolutions, too, since they rely on EffKAN. For this reason, for each KAN-based convolutional network we tested whether it was more advantageous to fix or train these parameters, both in KAN convolutions and in KAN linear layers.

For the preliminary study on KAN-based and traditional convolutional neural networks, four architectures are compared:
\begin{itemize}
    \item \textit{ConvNet}: a traditional convolutional network with two convolutional layers (5 $3 \times 3$ filters each) followed by two fully-connected layers (161 and 10 neurons);
    \item \textit{KANvNet}: a ConvNet in which the convolutional layers are replaced by CKAN layers (5 $3 \times 3$ filters each, with grid size set to 5 and spline order set to 3);
    \item \textit{ConvKANNet}: a ConvNet in which the fully-connected layers are replaced by EffKAN layers (20 and 10 edges);
    \item \textit{KKANNet}: a fully KAN-based architecture in which the feature extractor uses two CKAN layers (5 $3 \times 3$ filters each, grid size=5 and spline order=3) and the classifier has two EffKAN layers (31 and 10 edges).
\end{itemize}

The first architecture is a traditional convolutional network, \textit{ConvNet}. The feature extractor is made of two traditional convolutions (5 $3 \times 3$ filters), padded and activated by ReLU. The output of the feature extractor is down-sampled (max pooling with factor 2) and flattened before it is processed by a series of two linear layers: a ReLU-activated 980-to-161 layer and a softmax-activated 161-to-10 layer. This architecture has about 159k trainable parameters (in line with equations \ref{eq:paramConv_app} for the convolutions and \ref{eq:paramMLP_app} for the linear layers). \textit{KANvNet} replaces the convolutional layers in ConvNet with CKAN layers (5 $3 \times 3$ filters each, with grid size set to 5 and spline order set to 3), thus needing about 161.7k trainable parameters (considering equations \ref{eq:paramKConvFix_app} or \ref{eq:paramKConvTrain_app} for the feature extractor and equation \ref{eq:paramMLP_app} for the classifier). The choice of whether to fix or to optimize bias and scale factors mildly affects the total amount of trainable parameters, but the change is overall negligible with respect to the order of magnitude (163.79 vs 163.61 for KANvNet). \textit{ConvKANNet} replaces the linear layers in ConvNet with EffKAN layers (with 20 and 10 edges, grid size=5, and spline order=3), and requires about 159 trainable parameters (as per equations \ref{eq:paramConv_app} and \ref{eq:paramKANCL_app}). Finally, \textit{KKANNet} is fully KAN-based and replaces each traditional operation in ConvNet with KAN-based operations: the feature extractor is made by two 5-filter CKAN ($3 \times 3$, grid size=5, and spline order=3), while the has two EffKAN layers (31 and 10 edges, grid size=5, and spline order=3). KKANNet needs 158.2 trainable parameters (as per equations \ref{eq:paramKConvFix_app}-\ref{eq:paramKConvTrain_app} and \ref{eq:paramKANCL_app}).

\subsection{Results}
\label{subapp:resultsKANv}
\begin{figure}
    \centering
    \includegraphics[width=1\linewidth]{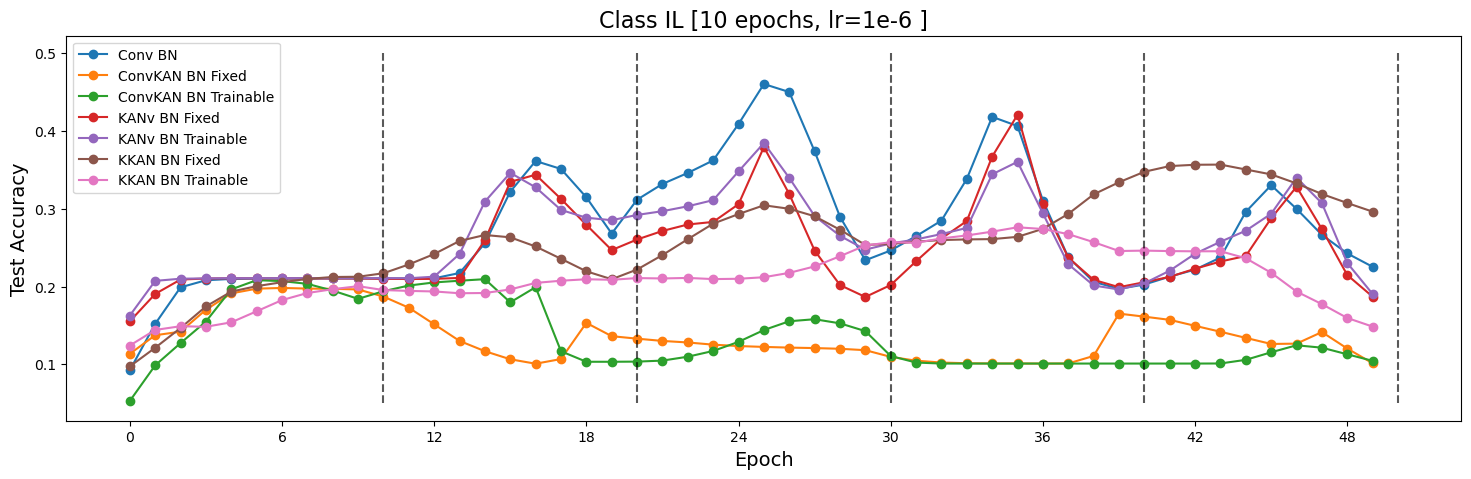}
    \caption{Test accuracy plots for the convolutional networks involved in this study. \textit{Fixed} means that the scale weights in the KAN layers were fixed; \textit{Trainable} means that they were optimized during training. \textit{BN} stands for batch normalization, that was added after any convolutional layer regardless of its nature.}
    \label{fig:plots_convNets}
\end{figure}

\begin{figure}
    \centering
    \centering
    \begin{minipage}[b]{0.30\textwidth}
        \centering
        \includegraphics[width=\linewidth, trim=1cm 1cm 1.8cm 1.5cm, clip]{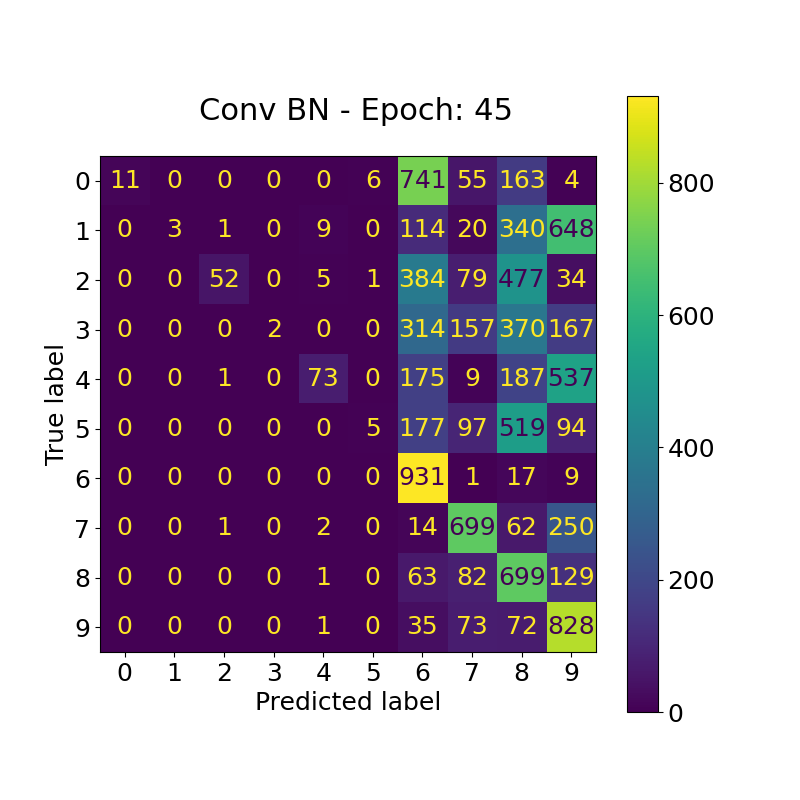}
        \subcaption{End of task 1.}\label{fig:confmatrConv_ep45}
    \end{minipage}
    \begin{minipage}[b]{0.30\textwidth}
        \centering
        \includegraphics[width=\linewidth, trim=1cm 1cm 1.8cm 1.5cm, clip]{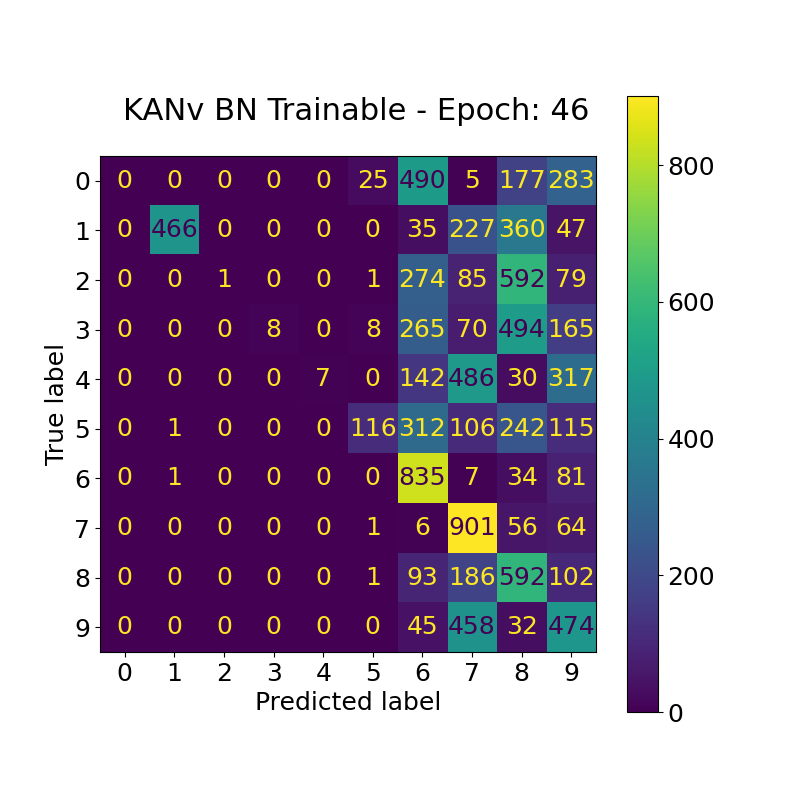}
        \subcaption{Third epoch of task 2.}\label{fig:confmatrKANv_ep46}
    \end{minipage}
    \begin{minipage}[b]{0.30\textwidth}
        \centering
        \includegraphics[width=\linewidth, trim=1cm 1cm 1.8cm 1.5cm, clip]{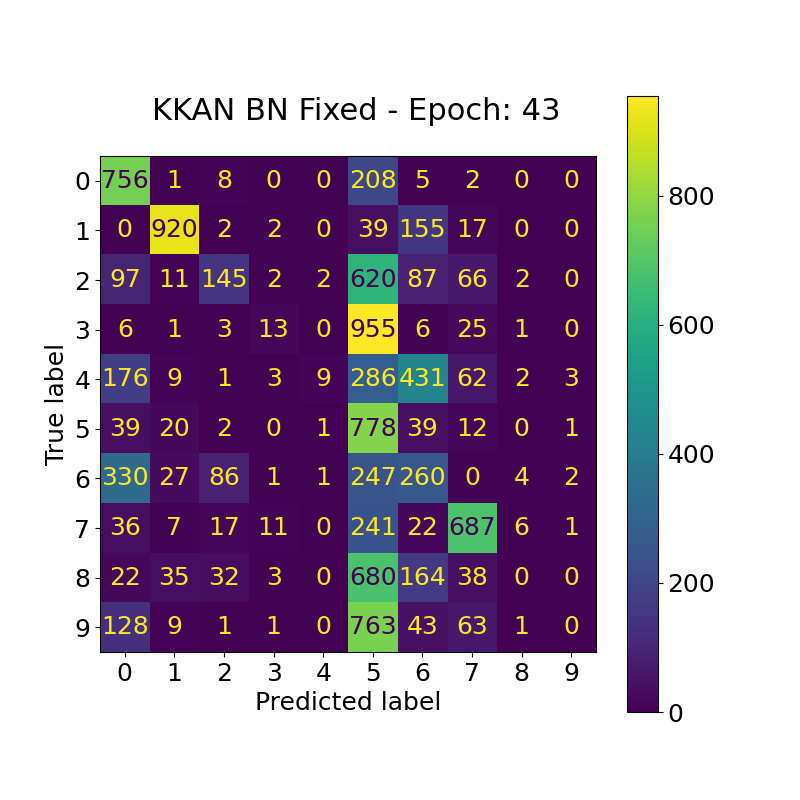}
        \subcaption{End of task 2.}\label{fig:confmatrKKAN_ep43}
    \end{minipage}
    \caption{Confusion matrices at the best respective epochs for ConvNet (BN), KANvNet (BN, trainable), and KKANNet (BN, fixed). }
    \label{fig:confmatrConvsKAN_best}
\end{figure}

Every network was trained on 50 epochs (10 epochs per task), with LR set to $10^-6$ and no LR decay. The training protocol defined in Sec. \ref{subsec:trainprot} was not applied for the exploration of convolutional networks, as these experiments serve primarily as a preliminary analysis of their capabilities. This study focused on evaluating the impact of scale factors $w_b$ and $w_s$ in KAN-based convolutions trained in CL settings. Consequently, for each of the three proposed KAN convolutional networks, we assessed whether fixing or training the scale weights in the KAN convolutional layers yielded better results. Additionally, batches were normalized after each convolution (both traditional and KAN-based) due to the poor performance exhibited by all models without batch normalization. Fig. \ref{fig:plots_convNets} shows the test accuracy trend for the convolutional and KAN-based convolutional networks involved in this preliminary study. In the legend, "BN" denotes batch normalization, while "Fixed" and "Trainable" indicate whether the scale factors were fixed or trained alongside the other parameters. All networks successfully learned the first task, achieving approximately 20\% test accuracy. However, the two ConvKAN models (comprising two traditional convolutions followed by two EffKAN linear layers) exhibited significant forgetting as soon as the second task begins, and show no signs of learning in the subsequent tasks. Overall, all the architectures displayed strong catastrophic forgetting and readily lost past information. This is confirmed by the confusion matrices, which we do not show here but can be found in our GitHub repository. In general, the highest test accuracy peaks are associated with confusion matrices that feature well-defined diagonals.

In the last task, the best performing network is \textit{KKANNet BN Fixed}, i.e., KKANNet in which the activations after each convolution are normalized and the scale factors are not optimized in KAN convolutions and KAN linear layers. However, ConvNet (BN) and KANvNet (BN, either fixed or trainable) achieve similar performance to KKANNet (BN, Fixed) in the last task. Interestingly, and unlike MLP, EffKAN, and PyKAN, no particular difference can be observed between the \say{Fixed} and \say{Trainable} versions of KANvNet. As shown in the confusion matrices reported in Fig. \ref{fig:confmatrConvsKAN_best}, when ConvNet and KANvNet achieve maximum test accuracy in task 5, they retain little to no information about the first three tasks, while showing moderate performance on the last two tasks. Conversely, when KKANNet achieves its peak performance in task 5, it has not learnt anything about the last task. In this sense, KKANNet behaves similarly to MLP and EffKAN (Figures \ref{fig:confmatrMLP} and \ref{fig:confmatrEffKAN}), the only architectures that were also completely unaware about the last task.  

%% file: KAN_CL.bbl
\begin{thebibliography}{22}
\providecommand{\natexlab}[1]{#1}
\providecommand{\url}[1]{\texttt{#1}}
\expandafter\ifx\csname urlstyle\endcsname\relax
  \providecommand{\doi}[1]{doi: #1}\else
  \providecommand{\doi}{doi: \begingroup \urlstyle{rm}\Url}\fi

\bibitem[Abraham and Robins(2005)]{abraham2005memory}
Wickliffe~C Abraham and Anthony Robins.
\newblock Memory retention--the synaptic stability versus plasticity dilemma.
\newblock \emph{Trends in neurosciences}, 28\penalty0 (2):\penalty0 73--78, 2005.

\bibitem[Blealtan(2024)]{efficient-kan}
Blealtan.
\newblock An efficient implementation of kolmogorov-arnold network, 2024.
\newblock URL \url{https://github.com/Blealtan/efficient-kan/tree/master}.

\bibitem[Bodner et~al.(2024)Bodner, Tepsich, Spolski, and Pourteau]{bodner2024ckan}
Alexander~Dylan Bodner, Antonio~Santiago Tepsich, Jack~Natan Spolski, and Santiago Pourteau.
\newblock Convolutional kolmogorov-arnold networks.
\newblock \emph{arXiv preprint arXiv:2406.13155}, 2024.

\bibitem[Bohra et~al.(2020)Bohra, Campos, Gupta, Aziznejad, and Unser]{bohra2020learning}
Pakshal Bohra, Joaquim Campos, Harshit Gupta, Shayan Aziznejad, and Michael Unser.
\newblock Learning activation functions in deep (spline) neural networks.
\newblock \emph{IEEE Open Journal of Signal Processing}, 1:\penalty0 295--309, 2020.

\bibitem[De~Boor(1978)]{de1978practical}
C~De~Boor.
\newblock A practical guide to splines.
\newblock \emph{Springer-Verlag google schola}, 2:\penalty0 4135--4195, 1978.

\bibitem[Dhiman(2024)]{dhiman2024kan}
Vikas Dhiman.
\newblock Kan: Kolmogorov--arnold networks: A review.
\newblock 2024.

\bibitem[Hornik et~al.(1989)Hornik, Stinchcombe, and White]{hornik1989multilayer}
Kurt Hornik, Maxwell Stinchcombe, and Halbert White.
\newblock Multilayer feedforward networks are universal approximators.
\newblock \emph{Neural networks}, 2\penalty0 (5):\penalty0 359--366, 1989.

\bibitem[LeCun et~al.(1998)LeCun, Bottou, Bengio, and Haffner]{lecun1998gradient}
Yann LeCun, L{\'e}on Bottou, Yoshua Bengio, and Patrick Haffner.
\newblock Gradient-based learning applied to document recognition.
\newblock \emph{Proceedings of the IEEE}, 86\penalty0 (11):\penalty0 2278--2324, 1998.

\bibitem[Lin(2024)]{awesomeKAN}
Jinhui Lin.
\newblock Awesome kan(kolmogorov-arnold network).
\newblock \url{https://github.com/mintisan/awesome-kan}, 2024.

\bibitem[Liu(2024)]{pyKAN}
Ziming Liu.
\newblock Kolmogorov-arnold networks (kan)).
\newblock \url{https://github.com/KindXiaoming/pykan}, 2024.

\bibitem[Liu et~al.(2024)Liu, Wang, Vaidya, Ruehle, Halverson, Solja{\v{c}}i{\'c}, Hou, and Tegmark]{liu2024kan}
Ziming Liu, Yixuan Wang, Sachin Vaidya, Fabian Ruehle, James Halverson, Marin Solja{\v{c}}i{\'c}, Thomas~Y Hou, and Max Tegmark.
\newblock Kan: Kolmogorov-arnold networks.
\newblock \emph{arXiv preprint arXiv:2404.19756}, 2024.

\bibitem[Lorentz(1962)]{lorentz1962metric}
GG~Lorentz.
\newblock Metric entropy, widths, and superpositions of functions.
\newblock \emph{The American Mathematical Monthly}, 69\penalty0 (6):\penalty0 469--485, 1962.

\bibitem[Rebuffi et~al.(2017)Rebuffi, Kolesnikov, Sperl, and Lampert]{rebuffi2017icarl}
Sylvestre-Alvise Rebuffi, Alexander Kolesnikov, Georg Sperl, and Christoph~H Lampert.
\newblock icarl: Incremental classifier and representation learning.
\newblock In \emph{Proceedings of the IEEE conference on Computer Vision and Pattern Recognition}, pages 2001--2010, 2017.

\bibitem[Rolnick et~al.(2019)Rolnick, Ahuja, Schwarz, Lillicrap, and Wayne]{rolnick2019experience}
David Rolnick, Arun Ahuja, Jonathan Schwarz, Timothy Lillicrap, and Gregory Wayne.
\newblock Experience replay for continual learning.
\newblock \emph{Advances in neural information processing systems}, 32, 2019.

\bibitem[Sovrasov(2018-2024)]{ptflops}
Vladislav Sovrasov.
\newblock ptflops: a flops counting tool for neural networks in pytorch framework, 2018-2024.
\newblock URL \url{https://github.com/sovrasov/flops-counter.pytorch}.

\bibitem[Sprecher(1965)]{sprecher1965structure}
David~A Sprecher.
\newblock On the structure of continuous functions of several variables.
\newblock \emph{Transactions of the American Mathematical Society}, 115:\penalty0 340--355, 1965.

\bibitem[Tepsich(2024)]{ckan}
Antonio Tepsich.
\newblock Convolutional kolmogorov-arnold network (ckan), 2024.
\newblock URL \url{https://github.com/AntonioTepsich/Convolutional-KANs}.

\bibitem[Van~de Ven and Tolias(2019)]{van2019three}
Gido~M Van~de Ven and Andreas~S Tolias.
\newblock Three scenarios for continual learning.
\newblock \emph{arXiv preprint arXiv:1904.07734}, 2019.

\bibitem[Van~de Ven et~al.(2022)Van~de Ven, Tuytelaars, and Tolias]{van2022three}
Gido~M Van~de Ven, Tinne Tuytelaars, and Andreas~S Tolias.
\newblock Three types of incremental learning.
\newblock \emph{Nature Machine Intelligence}, 4\penalty0 (12):\penalty0 1185--1197, 2022.

\bibitem[Wang et~al.(2024)Wang, Zhang, Su, and Zhu]{wang2024comprehensive}
Liyuan Wang, Xingxing Zhang, Hang Su, and Jun Zhu.
\newblock A comprehensive survey of continual learning: theory, method and application.
\newblock \emph{IEEE Transactions on Pattern Analysis and Machine Intelligence}, 2024.

\bibitem[Yu et~al.(2024)Yu, Yu, and Wang]{yu2024kan}
Runpeng Yu, Weihao Yu, and Xinchao Wang.
\newblock Kan or mlp: A fairer comparison.
\newblock \emph{arXiv preprint arXiv:2407.16674}, 2024.

\bibitem[Zhou et~al.(2024)Zhou, Wang, Qi, Ye, Zhan, and Liu]{zhou2024class}
Da-Wei Zhou, Qi-Wei Wang, Zhi-Hong Qi, Han-Jia Ye, De-Chuan Zhan, and Ziwei Liu.
\newblock Class-incremental learning: A survey.
\newblock \emph{IEEE Transactions on Pattern Analysis and Machine Intelligence}, 2024.

\end{thebibliography}
